\documentclass[10pt]{article} 
\usepackage[accepted]{tmlr}

\usepackage{amsmath,amsfonts,bm}

\def\Figref#1{Figure~\ref{#1}}

\def\Secref#1{Section~\ref{#1}}

\def\eqref#1{equation~\ref{#1}}
\def\Eqref#1{Equation~\ref{#1}}

\def\1{\bm{1}}

\DeclareMathAlphabet{\mathsfit}{\encodingdefault}{\sfdefault}{m}{sl}
\SetMathAlphabet{\mathsfit}{bold}{\encodingdefault}{\sfdefault}{bx}{n}

\def\h{{\bf h}}

\def\oo{{\bf o}}

\def\S{{\bf S}}

\def\x{{\bf x}}

\def\z{{\bf z}}

\def\0{{\bf 0}}
\def\1{{\bf 1}}

\numberwithin{theorem}{section}
\numberwithin{lemma}{section}
\numberwithin{remark}{section}
\numberwithin{cor}{section}
\numberwithin{proposition}{section}
\numberwithin{definition}{section}

\newcommand{\tabref}[1]{Table~\ref{#1}}
\newcommand{\appref}[1]{Appendix~\ref{#1}}

\renewcommand{\tilde}{\widetilde}
\renewcommand{\hat}{\widehat}

\newcommand{\method}{COAD}
\newcommand{\EE}{\mathbb{E}}
\newcommand{\red}[1]{\textcolor{red}{#1}}

\usepackage{hyperref}
\usepackage{url}
\usepackage{wrapfig}
\usepackage{microtype}
\usepackage{graphicx}
\usepackage{booktabs} 
\usepackage{tabularray}
\usepackage{subcaption}

\usepackage{amsmath}
\usepackage{amssymb}
\usepackage{mathtools}
\usepackage{amsthm}
\usepackage{bbm}
\usepackage{multirow}
\usepackage{wrapfig}
\usepackage{enumitem}

\usepackage[utf8]{inputenc} 
\usepackage[T1]{fontenc}    
\usepackage{url}            
\usepackage{booktabs}       
\usepackage{amsfonts}       
\usepackage{nicefrac}       
\usepackage{microtype}      
\usepackage{xcolor}         
\usepackage[normalem]{ulem}
\usepackage{array}
\usepackage{rotating}

\usepackage{booktabs} 
\usepackage{tabularray}
\usepackage{subcaption}
\usepackage{placeins}

\title{Causal Decoding for Hallucination-Resistant Multimodal Large Language Models}

\author{Shiwei Tan\textsuperscript{1*},~
Hengyi Wang\textsuperscript{1*},~
Weiyi Qin\textsuperscript{1*},~
Qi Xu\textsuperscript{2*},~
Zhigang Hua\textsuperscript{2*},~
Hao Wang\textsuperscript{1} \\\\
\textsuperscript{*}{Equal Contribution},
\textsuperscript{1}{Rutgers University},
\textsuperscript{2}{Meta Ranking AI}
}

\def\month{February}  
\def\year{2026} 
\def\openreview{\url{https://openreview.net/forum?id=5Wb5c0FaCG}} 

\begin{document}

\maketitle

\begin{abstract}
Multimodal Large Language Models (MLLMs) deliver detailed responses on vision-language tasks, yet remain susceptible to object hallucination (introducing objects not present in the image), undermining reliability in practice. Prior efforts often rely on heuristic penalties, post-hoc correction, or generic decoding tweaks, which do not directly intervene in the mechanisms that trigger object hallucination and thus yield limited gains. 
To address this challenge, we propose a causal decoding framework that applies targeted causal interventions during generation to curb spurious object mentions. By reshaping the decoding dynamics to attenuate spurious dependencies, our approach reduces false object tokens while maintaining descriptive quality. Across captioning and QA benchmarks, our framework substantially lowers object hallucination rates and achieves state-of-the-art faithfulness without degrading overall output quality.
\end{abstract}

\section{Introduction}\label{sec:intro}

\begin{wrapfigure}{r}{0.63\textwidth}
  \vspace{-0.4cm}
  \centering
  \includegraphics[width=0.63\textwidth]{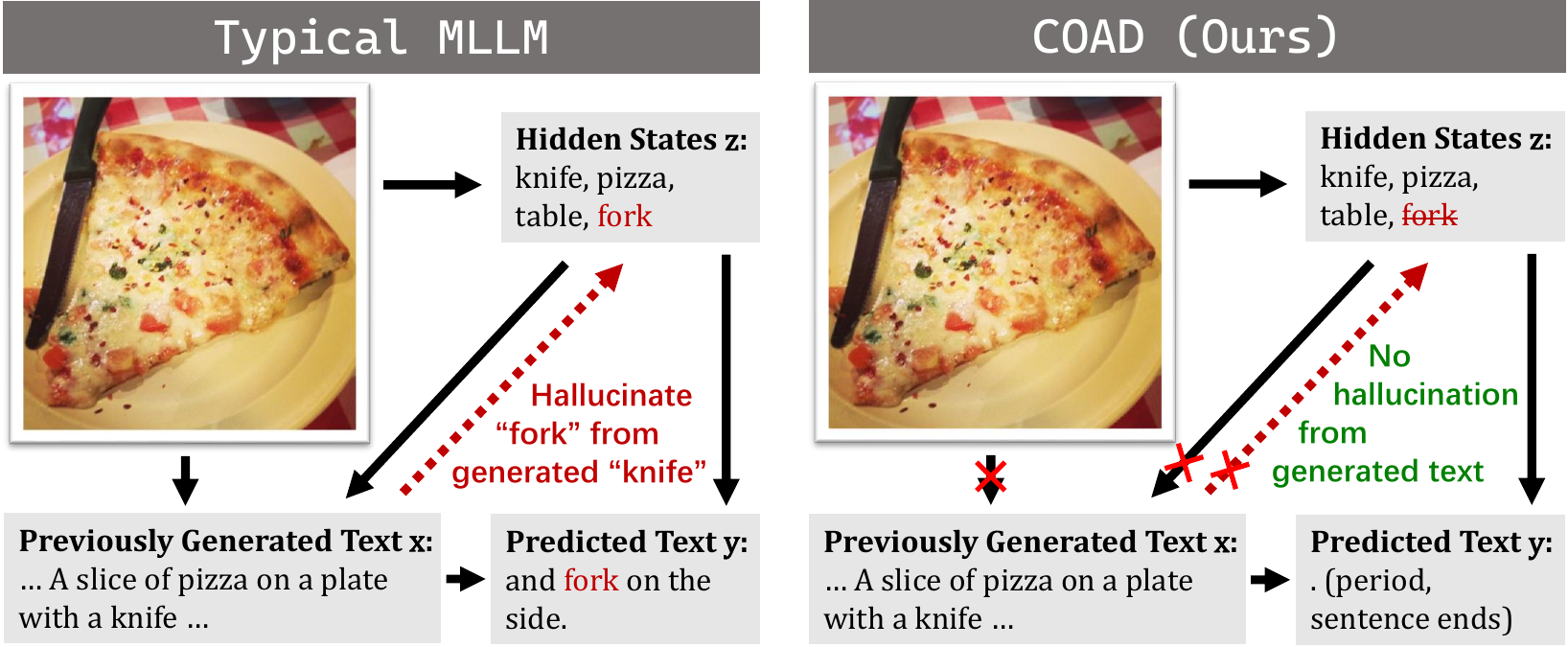}
  \vskip -0.2cm
  \caption{Simplified causal graphs for typical MLLMs and our COAD. \textbf{Left:} Typical MLLMs implicitly hallucinate objects (e.g., ``fork'') in the hidden states $\z$ due to previously generated text $\x$ (e.g., ``knife''). \textbf{Right:} Our COAD performs causal inference to remove links between the hidden states $\z$ and generated text $\x$, thereby avoiding hallucination.}
  \label{fig:teaser}
  \vskip -0.3cm
\end{wrapfigure}

Large language models (LLMs), such as GPT-4~\citep{gpt4} and LLaMA~\citep{llama}, have been rapidly developed and widely adopted due to their wide range of applications. To extend the capabilities of LLMs to visual tasks, multiple MLLMs have been proposed. Models such as LLaVA~\citep{llava} and MiniGPT~\citep{minigpt} typically project visual information into the same representational space as textual data, enabling a unified processing approach via an internal LLM. Although MLLMs have shown impressive performance in multimodal tasks, including chatbots, visual question answering, and image captioning, they remain susceptible to \emph{visual hallucination}.

Specifically, hallucinations in LLMs~\citep{llm-hall-survey} refer to cases where the model generates outputs that appear factual but are actually incorrect or ungrounded. With the introduction of visual inputs, multimodal LLMs (MLLMs) encounter a new category of hallucination: visual hallucination~\citep{lvlm-hall-survey}. Visual hallucination occurs when the MLLM output diverges from the content of the input image. This undermines the reliability of the models and restricts their applicability in high-stakes real-world scenarios that demand high precision, such as medical image analysis and legal document generation.

Recently, a variety of approaches have been proposed to mitigate hallucinations in MLLMs; they can be broadly categorized into \emph{two main strategies}: \textbf{(1)} The first strategy improves the model with external information, such as incorporating additional training data or retrieving knowledge from external source~\citep{lrv-instruction, hallucidoctor, recaption, dress, freshllms, rarr, stitchintime}. Although these methods effectively reduce hallucinations, they often require significant effort in data collection and depend on the quality and availability of external knowledge bases. \textbf{(2)} The second strategy aims to reduce hallucinations without relying on additional information, instead refining the training procedures of the model or improving the attention mechanisms during inference~\citep{eos, skipn, cad, vcd, pai, opera, dola, cgd, halc}. 
However, these methods still fail to model the causal effect from visual input (e.g., images) to the generated response. They are therefore often susceptible to confounding effect~\cite{CounTS,ICL,pearl2009causality} or bias brought by the generated text. As a result, they tend to generate new hallucinated text based on existing hallucinated text, exacerbating hallucination.

To address these challenges, we propose {Causal Object-Aware Decoding (COAD)} to reduce hallucination by incorporating causal inference into the model’s decoding process; this is inspired by hierarchical Bayesian deep learning~\cite{BDL,BDLSurvey,BLoB} and the causality literature~\cite{pearl2009causality,ICL,CounTS}. Specifically, we first employ an object detector to identify visual objects in the image, delegating part of the image comprehension task to this specialized component. We then expose these structured detection results to the MLLM by finetuning the MLLM with object detection outputs as additional inputs, alongside the image and previously generated text tokens. Finally, we perform causal inference to effectively integrate the predictions from both the original pretrained model and the finetuned model to generate the response.

COAD's design improves the reliability of the MLLM via enabling targeted interventions in the model’s understanding of visual objects. Furthermore, we incorporate causal inference to reduce the model’s dependence on self-generated text when processing and describing images, thereby promoting more stable and less hallucinatory outputs. 
Our contributions are as follows:
\begin{itemize}
\item We formulate the generation of reliable responses as the estimation of unknown oracle predictions and introduce a new framework, dubbed Causal Object-Aware Decoding (COAD), to reduce object hallucination.
\item We introduce a targeted intervention strategy that exposes and leverages visual structure, allowing the model to reason more faithfully about image content.
\item We provide empirical results to demonstrate the effectiveness of our method in improving generation quality and reducing object hallucination compared to state-of-the-art methods.
\end{itemize}

\section{Related Work}

\textbf{External Knowledge-Augmented Hallucination Mitigation.}
A typical strategy to mitigate hallucinations in MLLMs is to augment the model with external data. One line of work focuses on expanding or refining the training data to enhance grounding and reduce hallucinations~\citep{lrv-instruction, hallucidoctor, recaption, dress}. These methods typically involve curating high-quality multimodal instruction data, improving image-text alignment, or re-captioning visual content to ensure consistency with external world knowledge. By exposing the model to more reliable or better-aligned data, such approaches aim to reduce the risk of generating content that deviates from visual evidence or factual reality. Another line of research tackles hallucination at inference time by retrieving relevant information from external knowledge bases or the internet~\citep{freshllms, rarr, stitchintime}. These retrieval-augmented generation methods dynamically inject grounded knowledge into the model’s context, thereby improving factuality without requiring the model to memorize all details. 

While both approaches have demonstrated effectiveness, they rely on either significant data curation and annotation efforts or real-time access to high-quality and up-to-date external sources. In many real-world applications, especially those involving specialized or rapidly evolving domains, such requirements may not always be feasible or reliable, highlighting the need for alternative strategies that improve factual grounding without external dependencies.

\textbf{Internal Hallucination Mitigation.}
Other approaches mitigate hallucinations without relying on external data sources or retrieval mechanisms. These methods aim to improve the model’s internal decision-making process by modifying its behavior during training or inference. For instance, EOS~\citep{eos} encourages early stopping in sequence generation to prevent over-generation, which is often a source of factual inaccuracy. Skip-\textbackslash n~\citep{skipn} suppresses hallucinations by skipping newline tokens, which are empirically shown to precede low-quality or fabricated continuations. Several techniques reduce the distraction caused by noisy or misleading text-conditioned inputs by selectively emphasizing attention on visual tokens. Examples include CAD~\citep{cad}, VCD~\citep{vcd}, and PAI~\citep{pai}, which implement visual grounding and cross-modal alignment enhancements. CLIP-guided decoding~\citep{clip_guided} reduces hallucination by incorporating a CLIP-based image-text consistency score into a sentence-level beam search, adjusting the beam scores beyond the MLLM's own likelihood. OPERA~\citep{opera} proposes an intervention-based decoding strategy that penalizes overconfident token predictions, which are often associated with hallucinated content. DoLa~\citep{dola} improves factual alignment by comparing generation logits from early and late transformer layers, effectively regularizing token prediction based on layer-wise consistency. 
In this paper, we build on this line of research by focusing on internal mechanisms to reduce hallucination, without directly relying on external knowledge bases.


\section{Methodology}\label{sec:method}

In this section, we first introduce our COAD as a causal model for the MLLM's next-token generation process, and then describe how we apply causal inference to predict the next token during inference.

\subsection{Preliminaries and Key Intuition behind COAD}

\textbf{Problem Setting: Auto-Regressive Generation.}
We consider an auto-regressive MLLM that, at each decoding step, receives: (1) a model $M$; (2) an input image $\mathbf{S} \in \mathbb{R}^{c \times h \times w}$ where $c$ is the number of channels, $h$ is the height of the image, and $w$ is the width of the image; and (3) a sequence of previous tokens $\mathbf{x}$ (including the prompt and generated tokens so far). The model predicts the next token $y$ by sampling from
\[
y \sim P_{M}(y \mid \mathbf{x}, \mathbf{S}).
\]
Here $M$ may be a pretrained MLLM $M_p$, a finetuned model $M_f$, or a hypothetical oracle $M_\ast$ (introduced later).

\begin{wrapfigure}{r}{0.40\linewidth}
\vskip -0.55cm
\centering

\begin{subfigure}{0.48\linewidth}
    \centering
    \includegraphics[width=\linewidth]{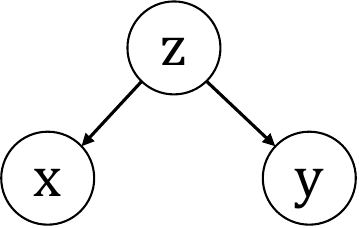}
    \caption{}
    \label{fig:example-graph-a}
\end{subfigure}
\hfill
\begin{subfigure}{0.48\linewidth}
    \centering
    \includegraphics[width=\linewidth]{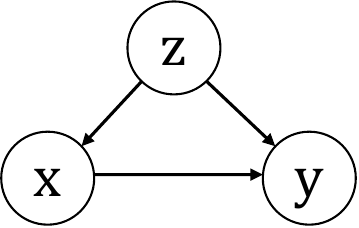}
    \caption{}
    \label{fig:example-graph-b}
\end{subfigure}

\vskip -0.2cm
\caption{\textbf{Illustration of confounding.}
(a) $z$ induces a spurious association between $x$ and $y$ even without a causal effect.
(b) Adding $x \rightarrow y$ introduces a genuine causal effect, but $P(y | x)$ remains confounded by $z$.}
\label{fig:example-graph}
\vskip -0.55cm
\end{wrapfigure}

\textbf{Causal Inference.}
Causal models provide a principled framework for distinguishing true causal effects from correlations induced by confounders~\citep{pearl2009causality}. A central tool is the use of interventions, denoted by $\mathrm{do}(\cdot)$, which remove spurious dependencies when analyzing the effect of one variable on another.

\textbf{Illustrative Example.} We next illustrate the effect of a confounder using the two causal graphs in \Figref{fig:example-graph}. Let $z$ denote temperature, $x$ the hot drink sales, and $y$ the ice cream sales.

In \Figref{fig:example-graph-a}, the edges reflect the causal structure where both $x$ and $y$ are influenced by the confounder $z$, but there is no direct causal effect from $x$ to $y$. In this setting, observing high hot drink sales $x$ implies that temperature $z$ is likely low, which in turn suggests that $y$ (ice cream sales) is also likely low. Consequently, even though $x$ does not causally affect $y$, the conditional probability $P(y | x)$ differs from $P(y)$. This discrepancy reflects a \emph{misleading spurious correlation} introduced by the confounder $z$.

To isolate the causal influence of $x$ on $y$, we instead compute the interventional distribution $P(y|\text{do}(x))$, which simulates actively setting $x$ to a fixed value while breaking its natural dependence on $z$. According to the rules of do-calculus, $P(y|\text{do}(x)) = P(y)$ in this case, correctly reflecting the absence of $x$'s causal influence on $y$. The formal derivation and rules can be found in~\citep{pearl2009causality}.

\Figref{fig:example-graph-b} then adds a direct causal edge $x \rightarrow y$ on top of \Figref{fig:example-graph-a}. In this case, hot drink sales $x$ indeed have a direct causal impact on ice cream sales $y$ (e.g., some customers may buy a cold item after consuming a hot drink). However, the conditional probability $P(y | x)$ still \emph{overestimates} this causal effect, because $z$ continues to influence both $x$ and $y$. The confounder $z$ thus causes $P(y | x)$ to capture both the genuine causal effect of $x$ on $y$ and the additional dependence mediated by $z$, motivating the use of interventional quantities such as $P(y | \mathrm{do}(x))$ to isolate the true causal influence.

\textbf{Analogy to MLLM Next-Token Prediction.}
The structure in \Figref{fig:example-graph-b} directly corresponds to MLLM decoding. Let $\x$ denote the previously generated tokens, $y$ the next token to be predicted, and $\z$ the model's hidden states representing its belief about which objects are present in the image (note that $\z$ is not the image itself, which could be modeled separately).
Because $\z$ is a backdoor variable connecting to both $\x$ and $y$, the conditional probability $P(y | \x)$ still \textbf{overestimates} the causal effect from $\x$ to $y$, i.e., overestimates the likelihood of certain tokens $y$, therefore potentially leading to hallucination.

\textbf{Key Intuition of COAD.}
To mitigate object hallucination, COAD explicitly models $\z$ as a variable representing object beliefs and replaces the standard conditional distribution with the interventional one $P(y | \mathrm{do}(\x), \z)$. This removes the spurious dependence introduced by the confounder $\z$ and predicts the next token $y$ based solely on the true causal effects of $\x$ and $\z$.

\textbf{Generative vs.\ Recognition Causal Models.}
The causal perspective adopted in this work follows the
\emph{recognition/inference} view commonly used in recent causal analyses of
vision-language models~\citep{geninterventionforcausal, causaltransforvr}.
Instead of modeling how the physical world generates an image (which would
typically imply a generative direction such as $\z \rightarrow \S$), our
objective is to describe how an MLLM processes an observed image $\S$ and
pre-existing text $\x$ in order to form internal beliefs and predict the next
token.

Under this recognition view, $\z$ denotes the model’s internal belief about
which objects are present in the input image, not a latent variable causing the
image itself. The information flow from $\S$ to $\z$ therefore represents the
model’s inference procedure, consistent with modern architectures such as LLaVA,
where visual features are computed by the vision encoder before any interaction
with textual tokens (see Appendix~\ref{app:llava-arch}). This viewpoint
provides the conceptual grounding for the causal structures introduced in the
following sections.

\subsection{Formal Definition of Object Hallucination}
\label{sec:formal-hallucination}

Let $\S$ be the input image, $\z^*$ the ground-truth set of visual objects, and 
$p_\theta(y | \x, \S)$ the model's predictive distribution for the next token $y$ given previous tokens $\x$ and the image $\S$. Let $p^*(y | \x, \z^*)$ denote the ``ground-truth'' conditional distribution, i.e., the distribution produced by an ideal model that fully respects the true visual semantics of the image.

We define object hallucination as the divergence between these two distributions:
\[
D\!\left(
p_\theta(y | \x, \S)
\;\|\;
p^*(y | \x, \z^*)
\right),
\]
where $D(\cdot\|\cdot)$ may be a KL or another divergence measure. A large divergence indicates that the model assigns high probability to tokens that contradict the true visual object content, thereby producing object hallucination.

In existing multimodal LLMs, hallucination often arises because the hidden states $\z$ may encode nonexistent objects based on previous tokens $\x$ rather than on the true image content $\S$. This can lead the model to generate tokens $y$ that are not visually grounded. We empirically validate this phenomenon for LLaVA using a linear-probe analysis of its internal object-existence beliefs; see Appendix~\ref{app:evidence} for details.

To address this issue, our COAD (i) uses a detector-derived proxy $\hat{\z}$ to approximate the visual constraints $\z^*$ in the ideal distribution $p^*(y | \x, \z^*)$, and (ii) intervenes on both the internal object-related hidden states $\z$ and the previous tokens $\x$ to block the spurious information pathway from $\x$ to $y$. These interventions reduce the divergence above and therefore mitigate object hallucination.

\subsection{Method Overview}

\begin{figure}[!t]
    \centering
    \includegraphics[width=1\linewidth]{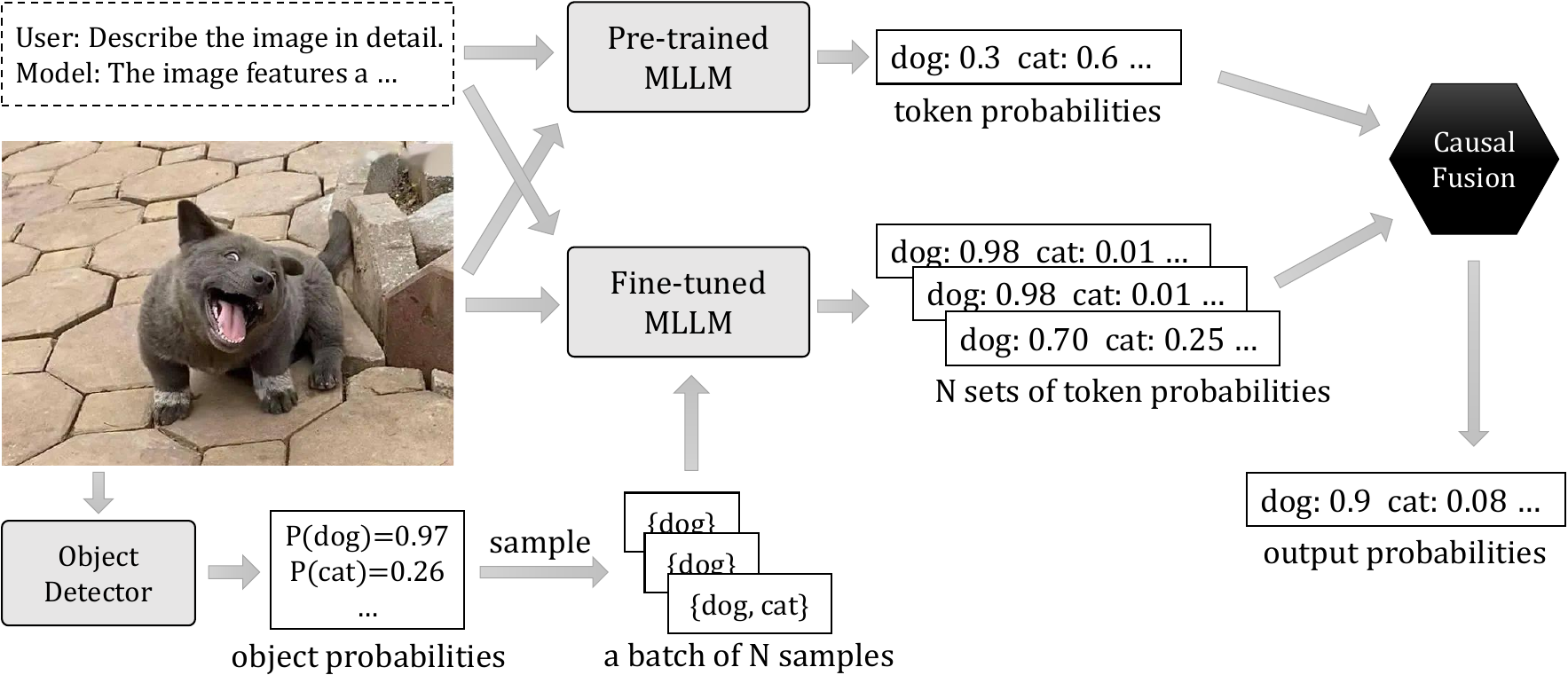}\vskip -0.1 cm
    \caption{\textbf{Overview of our COAD.} We employ an object detector to identify the objects present in an image. The MLLM is then finetuned to condition its token predictions on both these detected objects and the input image. COAD subsequently use causal inference to combine the output distributions of both the pretrained and finetuned MLLMs to generate the final prediction.}
    \label{fig:overview}
    \vskip -0.2cm
\end{figure}

\Figref{fig:overview} illustrates the decoding (i.e., text generation) process of our COAD. It assumes access to an MLLM that can incorporate a set of detected objects as additional context during generation. To achieve this, we finetune a pretrained MLLM with object-level information. During inference, an object detector identifies likely objects in the input image and outputs a probability distribution over candidate object classes. We then sample multiple plausible object sets from this distribution.

Each sampled object set is injected as an auxiliary input into the finetuned MLLM to produce a distribution over the next token. This results in $N$ next-token distributions, which are further combined with the distribution from the pretrained MLLM. Finally, COAD uses causal inference to combine these outputs to generate a more robust and object-aware prediction.

\subsection{Causal Model of a Standard MLLM}
\label{sec:mlm-causal-model}

Before introducing the causal model underlying \method{}, we first describe the
temporal causal structure of a standard MLLM during
autoregressive decoding. This view makes explicit how information flows across
timesteps and clarifies the relationship between the image, the evolving text
sequence, and the model's internal object-related variables.

\textbf{Temporal Structure.}
The input image $\S$ and the initial prompt $\x^{(0)}$ remain fixed throughout decoding.
The variable $\z$ represents the model's belief about which objects are present in the image and is determined solely by $\S$; it therefore remains constant across timesteps.
Here, $\z$ is a conceptual variable capturing image-conditioned object beliefs, rather than a token-level hidden state that evolves during decoding (see Appendix~\ref{app:implement} for further discussion).
This treatment is consistent with the architecture of modern vision-language models such as LLaVA, where visual features are produced by the vision encoder before any interaction with textual tokens (see Appendix~\ref{app:llava-arch}).
The only time-varying variables are the evolving text sequence $\x^{(t)}$ and the next token $y^{(t)}$.

\begin{figure}[t]
    \centering

    \begin{subfigure}[c]{0.48\linewidth}
        \centering
        \includegraphics[width=\linewidth]{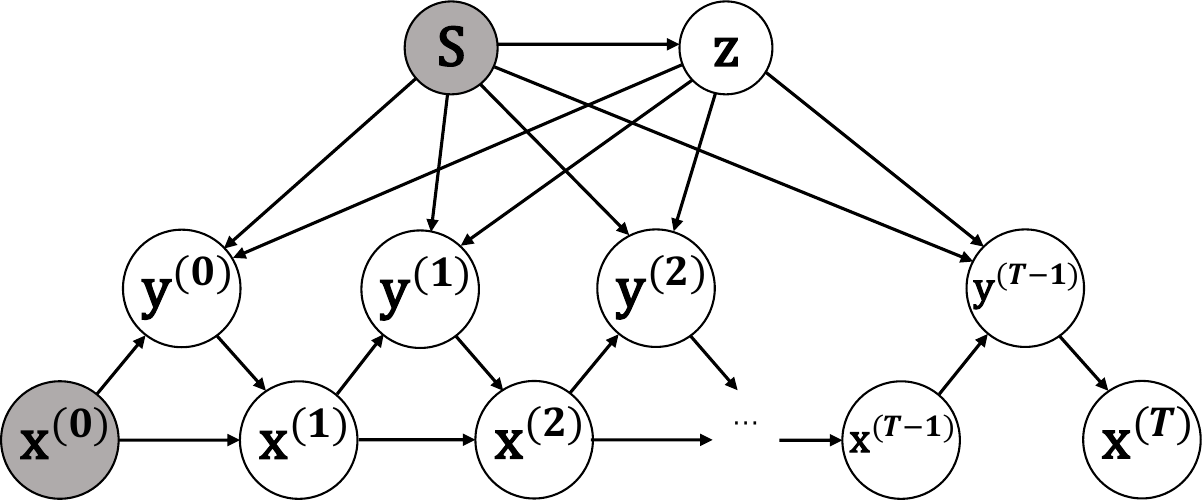}
        \subcaption{Full temporal causal graph of MLLM decoding process, where gray nodes represent observed variables.}
        \label{fig:origin-a}
    \end{subfigure}
    \hfill
    \begin{subfigure}[c]{0.48\linewidth}
        \centering

        \begin{subfigure}{\linewidth}
            \centering
            \includegraphics[width=\linewidth]{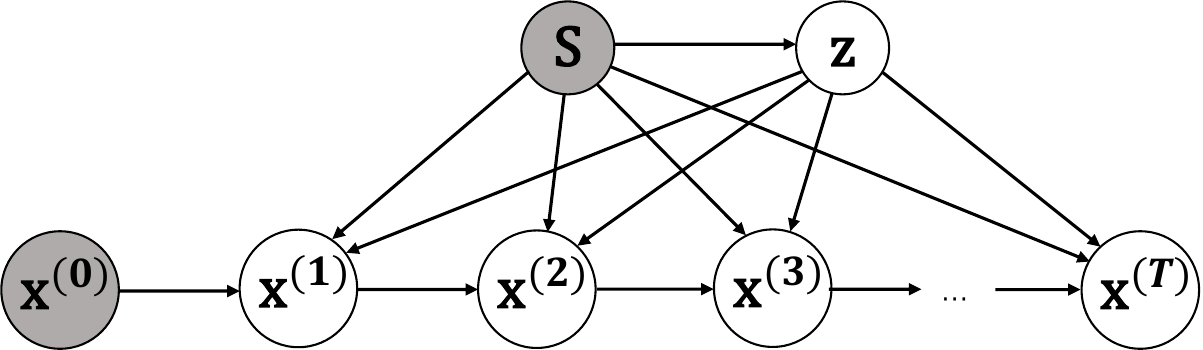}
            \subcaption{Collapsed version of (a).}
            \label{fig:origin-b}
        \end{subfigure}

        \vspace{0.8em}

        \begin{subfigure}{0.83\linewidth}
            \centering
            \includegraphics[width=0.6\linewidth]{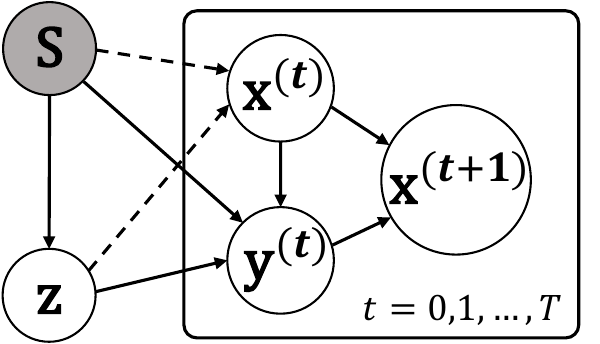}
            \subcaption{Time-compressed representation of (a).}
            \label{fig:origin-c}
        \end{subfigure}
    \end{subfigure}

    \caption{
        \textbf{Rolled-out causal structures of the MLLM decoding process.}
        \textbf{(a)} Full temporal (rolled-out) causal graph over decoding timesteps: the image $\S$ and initial prompt $\x^{(0)}$ are fixed, $\z$ denotes the image-derived object-belief variable (constant over time), and $\x^{(t)}$, $y^{(t)}$ evolve according to $\x^{(t)}, \S, \z \rightarrow y^{(t)}$ and $\x^{(t)}, y^{(t)} \rightarrow \x^{(t+1)}$.
        \textbf{(b)} Collapsed version of (a) obtained by treating $y^{(t)}$ as an implicit step in the autoregressive update, so that each $\x^{(t)}$ ($t>0$) is jointly determined by $\S$, $\z$, and $\x^{(t-1)}$.
        \textbf{(c)} Time-compressed representation in which the accumulated influence of all previous timesteps on $\x^{(t)}$ is summarized by dashed edges from $(\S, \z)$ to $\x^{(t)}$.
    }
    \label{fig:temporal_causal_graphs}
\end{figure}

\Figref{fig:origin-a} illustrates the resulting temporal
causal graph, which contains the following edges:
\begin{itemize}
    \item $\S \rightarrow \z$, indicating that the object-belief variable $\z$ is inferred from the image.
    \item $\x^{(t)}, \S, \z \rightarrow y^{(t)}$, reflecting that each predicted token depends on the current text, the visual input, and the inferred object beliefs.
    \item $\x^{(t)}, y^{(t)} \rightarrow \x^{(t+1)}$, reflecting the autoregressive update where each new token $y^{(t)}$ is appended to the old sequence $\x^{(t)}$ to form the new input sequence $\x^{(t+1)}$, i.e., the new input sequence is $\x^{(t+1)} = [\x^{(t)}, y^{(t)}].$
\end{itemize}

\textbf{Collapsed Representation.}
If the intermediate variables $y^{(t)}$ are treated as implicit steps in the autoregressive update, the subgraph of $(\x^{(t)}, y^{(t)}, \x^{(t+1)})$ can be collapsed into a single transition $\x^{(t)} \rightarrow \x^{(t+1)}$, producing the structure shown in \Figref{fig:origin-b}. In this form, each
$\x^{(t)}$ ($t>0$) is jointly determined by $\S$, $\z$, and $\x^{(t-1)}$.

\textbf{Time-Compressed View.}
Compressing the time axis of \Figref{fig:origin-a} yields the representation in \Figref{fig:origin-c}, which
summarizes the cumulative influence of past timesteps on $\x^{(t)}$ via dashed
edges from $(\S, \z)$ to $\x^{(t)}$. This compressed form closely matches the causal structure underlying a single decoding step, and provides
the foundation upon which we build the causal model for \method{} in the next
subsection.

\subsection{Causal Model of \method{}}\label{sec-causal-model}

\begin{figure}[!t]
  \centering
  \begin{subfigure}[b]{0.44\linewidth}
      \centering
      \includegraphics[width=\linewidth]{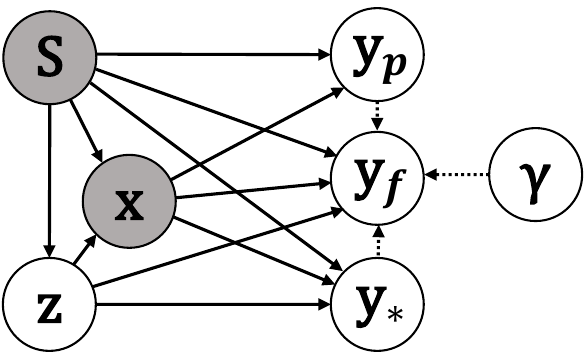}
      \caption{Original causal graph, where gray nodes represent observed variables. The dotted arrows illustrate an alternative pathway for generating $y_f$.}
      \label{fig:causal-a}
  \end{subfigure}
  \hfill
  \begin{subfigure}[b]{0.44\linewidth}
      \centering
      \includegraphics[width=\linewidth]{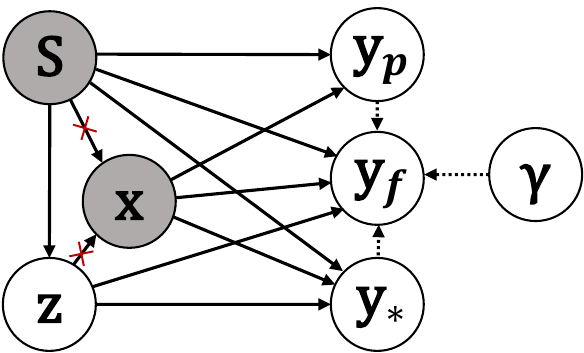}
      \caption{The resulting causal graph after $\text{do}(\x)$ intervention. All edges going into $\x$ are blocked by the intervention.}
      \label{fig:causal-b}
  \end{subfigure}
  \caption{Illustration of our COAD's causal model before and after intervention.}
  \label{fig:causal-both}
\end{figure}

\Figref{fig:causal-a} shows the causal model (as a causal Bayesian network) of our COAD. 
Given the input image $\S$ and the previous text tokens $\x$ as \emph{observed} variables, below are key components in COAD's generative process.

\textbf{Object Variable $\z$.} Our causal model operates at the granularity of individual token generation. At each decoding step, given the image $\S$ and the preceding (incomplete) text $\x$, COAD infers the presence of visual objects in $\S$ through a binary variable $\z \in \{0, 1\}^C$, where $C$ denotes the total number of object categories and is fixed by the choice of the detector. This variable is sampled from the distribution produced by an object detector $D$:
\begin{align*}
\z &\sim D(\S),
\end{align*}
where $D(\S) \in [0, 1]^C$ denotes the detector's estimated probability for the presence of each object category in the image.

\textbf{Dual MLLMs for Generation.} To model the next-token prediction, we incorporate two MLLMs into our causal framework: a pretrained model $M_p$ and a finetuned variant $M_f$. The pretrained model $M_p$ takes as input the image $\S$ and the preceding text $\x$, and outputs a distribution over the next token $y_p$. The finetuned model $M_f$, adapted from $M_p$, additionally conditions on the object variable $\z$ to produce a distribution over the next token $y_f$:
\begin{align*}
y_p &\sim P_{M_p}(y_p | \x, \S),\\
y_f &\sim P_{M_f}(y_f | \x, \S, \z),
\end{align*}
where $P_{M_p}(y_p | \x, \S)$ is the next token distribution predicted by $M_p$, and $P_{M_f}(y_f | \x, \S, \z)$ is the next token distribution predicted by $M_f$. In practice, $M_p$ and $M_f$ share most parameters for efficiency.

\textbf{Hypothetical Oracle MLLM.} To complete the causal graph, we introduce a hypothetical oracle model $M_*$, which serves as an idealized reference that always produces the optimal next-token distribution. The token predicted by this oracle, denoted as $y_*$, is generated as follows:
\begin{align*}
y_* &\sim P_{M_*}(y_* | \x, \S, \z),
\end{align*}
where $P_{M_*}(y_* | \x, \S, \z)$ represents the oracle's ground-truth distribution conditioned on the previous text $\x$, image $\S$, and object variable $\z$.

\textbf{Mixture-Based Generation.} We hypothesize that the finetuned model $M_f$ behaves as a mixture of the pretrained model $M_p$ and the hypothetical oracle model $M_*$. At each decoding step, $M_f$ may generate either the token predicted by $M_p$ or the one predicted by $M_*$, with a certain probability. Note that this is a natural assumption: $M_p$ serves as the initialization of $M_f$, and during finetuning, $M_f$ is optimized to better approximate ground-truth signals (as represented by $M_*$) while still inheriting behaviors from the original pretrained $M_p$.

To capture the uncertainty in $M_f$’s alignment between $M_p$ and $M_*$, we introduce a random variable $\gamma \in [0, 1]$, which governs the mixture proportion. This variable is drawn from a global prior Beta distribution with hyperparameters $\gamma_a, \gamma_b \in \mathbb{R}^+$, which are fixed across dataset:
\begin{align*}
\gamma &\sim \text{Beta}(\gamma_a, \gamma_b),
\end{align*}
and the next-token prediction $y_f$ is drawn approximately from a mixture of the two sources:
\begin{align*}
y_f &\approx \text{CategoricalMixture}(\{y_*, y_p\}, [\gamma, 1 - \gamma])\\
&\triangleq \text{Categorical}(\gamma\times y_*+ (1 - \gamma)\times y_p).
\end{align*}

This formulation reflects the intuition that $M_f$ may probabilistically interpolate between following the oracle model and reverting to its pretraining prior (more details in~\Eqref{eq:combined-model} below). It also provides an alternative generative view of $y_f$, which allows us to indirectly infer the oracle token $y_*$ in~\Secref{sec:inference}.

\textbf{Complete Causal Graph.} \Figref{fig:causal-a} summarizes the causal relationships among all random variables introduced in our model. The image $\S$ and previous tokens $\x$ are the only observed variables. Note that $\x$ itself may be influenced by $\S$ and $\z$ during previous decoding steps. All other variables are conditionally generated from their respective parents according to the mechanisms described above. The dotted connections from $y_*$, $y_p$, and $\gamma$ to $y_f$ indicate our hypothesis: $M_f$ can be alternatively interpreted as a probabilistic mixture of $M_*$ and $M_p$.

With the causal graph and the given observed variables, i.e., the image $\S$ and previous tokens $\x$, our goal is to (approximately) predict the oracle next token $y_*$ using causal inference. This will be discussed in~\Secref{sec:inference} below. 

\subsection{Inference Process}\label{sec:inference}
In this subsection, we describe how COAD employs our causal model to address the key challenges in reducing hallucinations of the MLLMs. We start by briefly discussing two key components of our method, i.e., \textbf{Causal Inference of Objects $\z$} and \textbf{Estimation of Oracle Predictions}, and then derive the corresponding equations that combine these two components.

\textbf{Component 1: Causal Inference of Objects $\z$.}
To ensure object beliefs reflect only the image content, we explicitly model them as variable $\z$ in our causal framework. Different from existing methods, where object belief is entangled in the hidden state and influenced by previous tokens $\x$, we block this dependency using an intervention $\text{do}(\x)$ (see \Figref{fig:causal-b}). This treats $\x$ as externally fixed, forcing the inference of $\z$ to depend solely on the image $\S$ and not on prior language outputs $\x$.

\textbf{Component 2: Estimation of Oracle Predictions.}
To approximate the oracle prediction $y_*$, we model the finetuned output $y_f$ as a mixture of the pretrained model $M_p$ and the oracle model $M_*$, following our assumption in \Secref{sec-causal-model}. While $y_*$ is unobservable, this mixture formulation allows us to estimate it using the available predictions $y_f$ and $y_p$ by $M_f$ and $M_p$, respectively. This provides a principled way to bridge the gap between observed model behavior and the ideal oracle output.

\textbf{Combining Components 1 \& 2 to Derive the Inference Objective.} By combining the previous components, our inference objective becomes computing the oracle prediction under intervention, i.e., $P(y_* | \S, \text{do}(\x))$. Using Bayes' rule and standard rules of causal inference~\citep{pearl2009causality}, we have that:
\begin{align}
  & P(y_*|\S,\text{do}(\x)) \label{eq:ultimate-goal}\\
= & \sum\nolimits_{\z} P(y_*|\S,\text{do}(\x),\z)P(\z|\S,\text{do}(\x)) \notag\\
= & \sum\nolimits_{\z} P(y_*|\S,\text{do}(\x),\z)P(\z|\S)\ &\text{(Rule 3)} \notag\\
= & \sum\nolimits_{\z} P(y_*|\S,\x,\z)P(\z|\S),\ &\text{(Rule 2)} \notag
\end{align}
This formulation rewrites the interventional query (with $\text{do}(\cdot)$) using standard conditional probabilities (without $\text{do}(\cdot)$), which can be estimated from observable components. We use the object detector $D$ to compute $P(\z | \S)$, which ensures that object beliefs are based solely on the image. The term $P(y_* | \S, \x, \z)$ represents the oracle model’s prediction, which is not directly accessible. To address this, we approximate it using a mixture model. Specifically, following our hypothesized relationship between $M_f$, $M_*$, and $M_p$, we have that:
\begin{align}
P(y_f|\S,\x,\z)=\EE_\gamma\big[\gamma P(y_*|\S,\x,\z)+ (1-\gamma)P(y_p|\S,\x)\big]. \label{eq:combined-model}
\end{align}
By rearranging~\Eqref{eq:combined-model}, we can rewrite the prediction from $M_*$ in terms of the predictions $y_p$ and $y_f$ from $M_p$ and $M_f$, respectively. Specifically:
\begin{align}
&P(y_*|\S,\x,\z) \label{eq:ystar-expand}\\
=&\tfrac{1}{\EE_\gamma[\gamma]}P(y_f|\S,\x,\z) + (1-\tfrac{1}{\EE_\gamma[\gamma]}) P(y_p|\S,\x) \notag \\
=&\left(1+\tfrac{\gamma_b}{\gamma_a}\right)P(y_f|\S,\x,\z) - \left(\tfrac{\gamma_b}{\gamma_a}\right) P(y_p|\S,\x). \notag
\end{align}
\textbf{Final Inference Objective.} After substituting~\Eqref{eq:ystar-expand} into~\Eqref{eq:ultimate-goal} and rearranging the terms, we can then rewrite our final inference objective as a combination of known quantities:
\begin{align}
&P(y_*|\S,\text{do}(\x)) \label{eq:final-formula}\\
=&\sum\nolimits_{\z} P(\z|\S)\big[ \left(1+\alpha\right)P(y_f|\S,\x,\z) - \alpha P(y_p|\S,\x)\big] \notag \\
=&\left(1+\alpha\right)\sum\nolimits_{\z} \big[P(\z|\S) P(y_f|\S,\x,\z)\big] - \alpha P(y_p|\S,\x). \notag
\end{align}
where we use the shorthand $\alpha \triangleq \gamma_b / \gamma_a$. Since only the ratio $\alpha = \gamma_b / \gamma_a$ appears in the final expression, the Beta distribution's parameters $(\gamma_a, \gamma_b)$ (as hyperparameters) effectively have only one degree of freedom during inference. In practice, We therefore treat $\alpha$ as a single global hyperparameter (see Appendix~\ref{app:implement} for additional details). Notably, computing the closed-form expression in~\Eqref{eq:final-formula} is equivalent to sampling many values of $\gamma \sim \mathrm{Beta}(\gamma_a,\gamma_b)$ and averaging their contributions in expectation. This closed-form solution makes COAD both more efficient and conceptually consistent with the underlying mixture interpretation.

Since the dimension of $\z$ can be large, directly summing over all possible
object-belief vectors is computationally intractable. We consider two practical implementations to approximate the expectation over $\z$ in \Eqref{eq:final-formula}: (1) \emph{Monte Carlo sampling}, where we treat the detector's output $\tilde{z}\in [0,1]^C$ as the parameters of $C$ Bernoulli distributions and sample $N$ binary vectors $z_i\in \{0,1\}^C$ (where $i=1,2,\dots,N$) from them, so that the term $\sum_{\z} P(\z|\S)P(y_f|\S,\x,\z)$ is approximated by $\frac{1}{N}\sum_{i=1}^N P(y_f|\S,\x,\z_i)$; and (2) \emph{probability-based approximation}, where we directly feed the probability vector $\tilde{\z}$ into $M_f$, which empirically provides an efficient approximation to the same expectation while avoiding sampling. Due to its high efficiency, we adopt the second implementation (although it is an approximation) for all our experiments.

\textbf{Summary of \method{}.}
To summarize, training and inference of \method{} consist of the following steps:

\begin{enumerate}
\item Modify the pretrained MLLM to accept an object belief vector $\z$ as an additional input.
\item Finetune the modified MLLM using the object vectors $\z$ (predicted by an object detector).
\item At inference time, compute the next-token probability using~\Eqref{eq:final-formula}, approximating the expectation over $\z$ via Monte Carlo sampling.
\end{enumerate}

Therefore, \method{} enables \emph{object-aware dehallucination} by explicitly grounding language generation in visual object beliefs. Through causal modeling and intervention, \method{}  ensures that predictions remain faithful to the image content, reducing reliance on spurious correlations from prior text.


\section{Experiments}\label{sec:exp}
In this section, we compare COAD with existing methods on real-world datasets.

\subsection{Datasets and Metrics}
We use various datasets and metrics below to evaluate the MLLMs.

\textbf{POPE.} The Polling-based Object Probing Evaluation (POPE)~\citep{li2023pope} employs visual question answering to assess whether an MLLM can correctly identify the presence of an object in an input image. Following the literature~\citep{pai,opera}, we focus on the MSCOCO dataset with 500 images, with each image having 6 questions for each split of POPE. We evaluate the object recognition performance using the Precision, Recall, F-1, and Accuracy metrics.

\textbf{CHAIR.} Caption Hallucination Assessment with Image Relevance (CHAIR)~\citep{rohrbach2018chair} is a set of widely used metrics to evaluate captioning hallucination. Following the literature~\citep{pai,opera}, we use the MSCOCO dataset~\citep{MSCOCO} that provides annotations for ground-truth objects in images. Specifically, CHAIR includes two metrics: 
    \begin{itemize}
        \item \textbf{CHAIR$_S$}, which measures the proportion of captions containing hallucinated objects relative to the total number of captions:
\begin{align*}
    \text{CHAIR}_S=\vert\text{captions with hallucinated objects}\vert\ \slash\ \vert\text{all captions}\vert,
\end{align*}
        \item \textbf{CHAIR$_I$}, which measures the proportion of the hallucinated objects relative to the total number of mentioned objects:
\begin{align*}
    \text{CHAIR}_I=\vert\text{hallucinated objects}\vert\ \slash\ \vert\text{all mentioned objects}\vert.
\end{align*}
    \end{itemize}

\textbf{MMHal-Bench.} MMHal-Bench~\citep{sun2023mmhal} is a dataset designed to evaluate MLLMs on diverse questions where they may produce false claims about image content. The benchmark spans eight hallucination dimensions that assess different aspects of visual grounding: 
(1) \textit{Object attribute}: the ability to correctly describe properties such as color or shape; 
(2) \textit{Adversarial object}: the ability to recognize when a queried object is not present rather than hallucinating it; 
(3) \textit{Comparison}: the ability to compare attributes or properties across multiple objects; 
(4) \textit{Counting}: the ability to estimate the number of referred objects; 
(5) \textit{Spatial relation}: the ability to reason about relative positions among objects; 
(6) \textit{Environment}: the ability to infer aspects of the surrounding scene or background; 
(7) \textit{Holistic description}: the ability to provide accurate global descriptions of the entire image; 
(8) \textit{Others}: the ability to recognize text or symbols and reason based on observable visual information. 
Scores for each dimension are computed as the proportion of responses deemed non-hallucinatory. Higher scores across these dimensions indicate better grounding and fewer hallucinations. Following the MMHal-Bench evaluation protocol~\citep{sun2023mmhal}, we use GPT-4 to perform this judgment.

\subsection{Baselines}
We use LLaVA-1.5-7B~\citep{llava} as the base model for all evaluated methods. For \method{}, we use RTMDet~\citep{rtmdet} as the object detector $D$.

As discussed in \Secref{sec:intro}, existing hallucination-mitigation approaches fall into two broad categories: (1) external-knowledge-based methods and (2) internal architecture/decoding modifications. Since \method{} belongs to the second family and does not rely on external data retrieval, all baselines evaluated here are drawn from this internal-mechanism category to ensure a fair comparison. Under this setting, we compare \method{} with state-of-the-art internal-mechanism methods, including 
Decoding by Contrasting Layers (\textbf{DoLa})~\citep{dola}, 
Paying More Attention to Image (\textbf{PAI})~\citep{pai}, 
End-of-Sentence Decision (\textbf{EOS})~\citep{eos}, 
Over-trust Penalty and Retrospection-Allocation (\textbf{OPERA})~\citep{opera}, 
Visual Contrastive Decoding (\textbf{VCD})~\citep{vcd}, 
Context-Aware Decoding (\textbf{CAD})~\citep{cad}, 
and Object Hallucination Reduction via Adaptive Focal-Contrast Decoding (\textbf{HALC})~\citep{halc}.

\subsection{Implementation Details}

We finetune \method{} on a subset of MSCOCO images sourced from the LLaVA dataset. To enable the model to incorporate the auxiliary input $\z$, we introduce a two-layer MLP projector (with hidden size $256$) that maps $\z$ into the token embedding space, following LLaVA’s multimodal token integration approach. We employ LoRA ($r=128$, $\alpha=256$), a cosine learning rate schedule with an initial learning rate of $4\text{e}{-5}$, a batch size of 128, and train the model for 1 epoch. 
During inference, we use sampling by default with temperature 0.2 and a maximum of 512 output tokens. See~\appref{app:implement} for more details.

\subsection{Main Results}\label{sec:results}

In this section, we compare \method{} with different baselines across various datasets and metrics.

\begin{figure}[!t]
    \centering
    \includegraphics[width=1\linewidth]{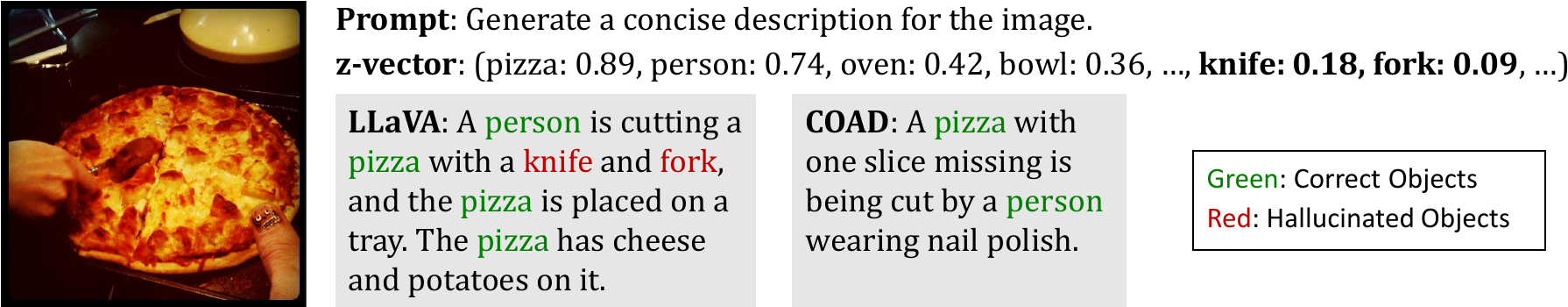}\vskip -0.1 cm
    \caption{\textbf{Case study on caption generation.} MSCOCO objects mentioned in the text are highlighted in \textcolor[RGB]{192,0,0}{red} (hallucinated) or \textcolor[RGB]{0,128,0}{green} (correct). We compare the baseline LLaVA with our COAD-enhanced model. While LLaVA hallucinates nonexistent objects (e.g., \textbf{knife} and \textbf{fork}), the $\z$-vector produced by the object detector suggests that these objects are absent. By leveraging this signal, COAD produces a faithful caption grounded in the actual image content, consistent with the improvements shown in CHAIR metrics.}
    \label{fig:case_study}
    \vskip -0.2cm
\end{figure}

\begin{table}[!t]
\centering
\caption{Comparison of different methods in terms of CHAIR metrics. \textbf{Boldface} and \underline{underlining} denote the best and the second-best performance, respectively.}
\renewcommand{\arraystretch}{1.2}
\label{tab:chair}
\vskip -0.2cm
\begin{tabular*}{1\linewidth}{@{\extracolsep{\fill}}lccccccccc}
\toprule
\textbf{Method} & Base & PAI & DoLa & VCD & CAD & OPERA & EOS & HALC & \method{} \\
\midrule
$\text{CHAIR}_I\downarrow$ & 9.9 & 5.8 & 13.0 & 11.4 & 9.9 & \uline{4.5} & 5.8 & 5.2 & \textbf{3.4} \\
$\text{CHAIR}_S\downarrow$ & 29.6 & 11.3 & 37.0 & 32.5 & 28.0 & \uline{7.4} & 10.6 & 11.1 & \textbf{5.3} \\
\bottomrule
\end{tabular*}
\vskip -0.2cm
\end{table}

\textbf{Free-Form Generation Evaluation on CHAIR.} 
We first evaluate \method{} on the CHAIR benchmark, which measures hallucination rates in free-form image captioning. The CHAIR benchmark includes two sub-metrics: CHAIR$_I$ (instance-level) and CHAIR$_S$ (sentence-level). Lower CHAIR$_I$ and CHAIR$_S$ indicate fewer hallucinated mentions. 

As shown in~\tabref{tab:chair}, \method{} achieves the best performance across all three CHAIR metrics, significantly reducing hallucinations. Specifically, it achieves 3.4 and 5.3 in terms of CHAIR$_I$ and CHAIR$_S$, respectively, outperforming all existing baselines. This demonstrates that our causal object-aware decoding effectively reduces hallucination of generated captions.

\Figref{fig:case_study} shows a qualitative example comparing the baseline LLaVA and our COAD. Here, LLaVA hallucinates nonexistent objects such as a \textit{knife} and \textit{fork}, while our \method{} correctly suppresses them and generates a more faithful caption.\footnote{While \method{} successfully removes the hallucinated objects such as the ``knife'' and ``fork'', its output also includes the phrase ``one slice missing'', which may be an inaccurate description of the pizza. This type of error concerns the \emph{attributes} of an already-present object rather than the presence of additional objects. Since \method{} is designed specifically to mitigate \emph{object hallucination}, such attribute-level inconsistencies fall outside the scope of what our method targets.} This illustrates how causal object-aware decoding helps mitigate hallucination in practice. Additional case studies are provided in \appref{app:case_study}.

\begin{table}[!t]
\centering
\setlength{\tabcolsep}{3pt}
\renewcommand{\arraystretch}{1.2}
\caption{Evaluation on MMHal-Bench across 8 hallucination dimensions: attributes (attr), adversarial objects (adv), comparison (cmp), counting (cnt), spatial relations (rel), environment (env), holistic/overall description (hol), and others (oth). \textbf{Boldface} and \underline{underlining} denote the best and the second-best performance, respectively.}
\vskip -0.2cm
\label{tab:mmhal}
\begin{tabular*}{1\linewidth}{l@{\extracolsep{\fill}}cccccccccc}
\toprule
\textbf{Method} & \textbf{Avg.~Score} & \textbf{Hall.~Rate} 
& \textbf{attr} & \textbf{adv} & \textbf{cmp} & \textbf{cnt} & \textbf{rel} & \textbf{env} & \textbf{hol} & \textbf{oth} \\
\midrule
Base      & 1.88 & 0.68 & 2.33 & 1.25 & 2.67 & 0.83 & 1.75 & 3.17 & 1.42 & 1.58 \\
PAI       & 2.10 & 0.65 & 1.92 & 1.33 & 2.25 & 2.17 & \uline{2.17} & 3.67 & \uline{1.75} & 1.58 \\
Dola      & 2.01 & \uline{0.62} & 2.08 & 1.42 & 2.75 & 1.67 & 1.17 & \textbf{4.00} & \uline{1.75} & 1.25 \\
VCD       & 1.98 & 0.67 & 2.17 & \textbf{1.83} & 1.83 & 1.33 & \textbf{2.42} & 3.33 & 1.33 & 1.58 \\
CAD       & 2.00 & 0.64 & 2.50 & 1.25 & 2.42 & 0.75 & 1.33 & \uline{3.83} & \textbf{1.83} & \uline{2.08} \\
OPERA     & 2.09 & 0.65 & 2.58 & \uline{1.67} & 2.67 & \textbf{2.50} & 1.58 & 3.08 & 1.17 & 1.50 \\
EOS       & 2.08 & \uline{0.62} & \uline{2.67} & 1.33 & 2.67 & 1.00 & 1.83 & 3.17 & 1.58 & \textbf{2.42} \\
HALC      & \uline{2.12} & 0.64 & 2.33 & \uline{1.67} & \uline{3.00} & \uline{2.25} & 1.67 & 3.42 & 1.33 & 1.33 \\
\method{} & \textbf{2.52} & \textbf{0.52} & \textbf{3.58} & \textbf{1.83} & \textbf{3.33} & 2.08 & 2.08 & 3.50 & 1.33 & \textbf{2.42} \\
\bottomrule
\end{tabular*}
\vskip -0.4cm
\end{table}

\textbf{Multimodal QA Evaluation on MMHal-Bench.} 
\tabref{tab:mmhal} shows the results on MMHal-Bench. \method{} achieves the highest average score (2.52) and the lowest hallucination rate (0.52), significantly outperforming all baselines. The strong performance is consistent across multiple benchmark subsets, particularly in the Attribute, Comparison, and Relation categories, indicating improved factual accuracy and reasoning. These results further demonstrate that incorporating object-level cues effectively reduces hallucination while maintaining or enhancing generation quality.

\begin{table}[!t]
\centering
\setlength{\tabcolsep}{3pt}
\renewcommand{\arraystretch}{1.2}
\caption{POPE evaluation results on the MSCOCO dataset. \textbf{Boldface} and \underline{underlining} denote the best and the second-best performance, respectively.}
\label{tab:pope}
\vskip -0.2cm
\begin{tabular*}{1\linewidth}{l@{\extracolsep{\fill}} cccc c cccc c cccc c}
\toprule
\multirow{2}{*}{\textbf{Method}} 
  & \multicolumn{5}{c}{\textbf{Random}} \hspace{6pt} 
  & \multicolumn{5}{c}{\textbf{Popular}} \hspace{6pt} 
  & \multicolumn{5}{c}{\textbf{Adversarial}} \\
\cmidrule(lr){2-6}\cmidrule(lr){7-11}\cmidrule(lr){12-16}
 & Acc & P & R & F1 & Yes 
 & Acc & P & R & F1 & Yes 
 & Acc & P & R & F1 & Yes \\
\midrule
Base      & 89.0 & 89.3 & 88.6 & 89.0 & 49.6 & 85.0 & 82.6 & 88.7 & 85.5 & 53.7 & 78.8 & 74.0 & 88.8 & 80.8 & 60.0 \\
PAI       & \uline{89.3} & 89.6 & 88.9 & \uline{89.2} & 49.6 & \textbf{86.1} & \uline{84.2} & \uline{89.0} & \textbf{86.5} & 52.9 & 78.9 & 74.4 & 88.3 & 80.7 & 59.4 \\
Dola      & 86.3 & 85.3 & 87.7 & 86.5 & 51.4 & 83.0 & 80.5 & 87.1 & 83.6 & 54.1 & 78.2 & 73.9 & 87.4 & 80.1 & 59.2 \\
VCD       & 88.8 & 88.8 & 88.7 & 88.8 & 50.0 & 85.4 & 83.6 & 88.1 & 85.8 & 52.7 & \uline{79.2} & 74.5 & 88.6 & \uline{81.0} & 59.4 \\
CAD       & 88.6 & 88.7 & 88.5 & 88.6 & 49.9 & 84.8 & 82.5 & 88.3 & 85.3 & 53.5 & 78.5 & 74.0 & 87.9 & 80.3 & 59.4 \\
OPERA     & \textbf{89.4} & \uline{89.7} & \uline{89.0} & \textbf{89.3} & 49.6 & \uline{85.9} & 83.9 & \uline{89.0} & \uline{86.4} & 53.1 & 79.1 & 74.3 & 89.0 & \uline{81.0} & 59.9 \\
EOS       & 85.4 & 81.5 & \textbf{91.7} & 86.3 & 56.3 & 81.2 & 75.8 & \textbf{91.7} & 83.0 & 60.5 & 75.9 & 69.6 & \textbf{91.9} & 79.2 & 66.0 \\
HALC      & 88.7 & \textbf{89.9} & 87.1 & 88.5 & 48.5 & 85.8 & \textbf{84.8} & 87.1 & 86.0 & 51.4 & 79.1 & \uline{75.0} & 87.1 & 80.6 & 58.1 \\
\method{} & 89.0 & 89.6 & 88.3 & 89.0 & 49.3 & 85.5 & 84.0 & 87.6 & 85.8 & 52.1 & \textbf{79.8} & \textbf{75.8} & 87.5 & \textbf{81.2} & 57.7 \\
\bottomrule
\end{tabular*}
\vskip -0.2cm
\end{table}

\textbf{Object Probing Evaluation on POPE.} 
\tabref{tab:pope} shows the POPE evaluation results across three settings. \method{} achieves the highest accuracy (79.8) and F1 score (81.2) on the Adversarial subset, outperforming all baselines, indicating better robustness to prompts designed to induce hallucination. In the Popular and Random subsets, it performs comparably to state-of-the-art methods in F1 while maintaining a low hallucination ratio. These results confirm that our approach effectively reduces hallucinations while preserving factual precision across diverse input types.

\begin{wraptable}{r}{0.35\linewidth}
\centering
\vskip -0.55cm
\caption{Results of \method{} and ablations on CHAIR. ``$M_f$ only'' means only using the finetuned model $M_f$ for generation; ``w/o $\z$'' means replacing $M_f$ by a normally finetuned MLLM and applying \method{}, without any $\z$ vectors involved in the whole process.} 
\label{tab:ablation}
\vskip -0.25cm
\renewcommand{\arraystretch}{1.2}
\resizebox{\linewidth}{!}{%
\begin{tabular}{@{\hskip 1pt}l@{\hskip -1pt}c@{\hskip 5pt}c@{\hskip 1pt}}
\toprule
\textbf{Method}        & CHAIR$_I \downarrow$ & CHAIR$_S \downarrow$ \\
\midrule
\method{} (Full)     & \textbf{3.4} & \textbf{5.3} \\
\method{} ($M_f$ only)    & 5.4 & 10.8 \\
\method{} (w/o $\mathbf{z}$) & 6.9 & 18.1 \\
\bottomrule
\end{tabular}
}
\vskip -1.5cm
\end{wraptable}

\textbf{Ablation Studies.} 
We conduct two ablation studies on \textsc{CHAIR} to better understand the source of our improvements. 
Specifically, we compare our full COAD with (1) ``\textbf{COAD ($M_f$ Only)}'', which only uses the finetuned model $M_f$ without applying our causal decoding procedure and (2) ``\textbf{COAD (w/o $\z$)}'', where we train $M_f$ without $\z$ and perform causal decoding using this modified $M_f$. \tabref{tab:ablation} shows the results. The gap between COAD and ``{COAD ($M_f$ Only)}'' verifies the effectiveness of our causal decoding algorithm, while the gap between COAD and ``{COAD (w/o $\z$)}'' verifies the important role of $\z$ in COAD (see more discussion in~\appref{app:ablation}).

\subsection{Runtime and Computational Overhead}
\label{sec:runtime}

Beyond hallucination metrics, we also compare the computational overhead of \method{} with existing methods. To ensure consistency with the CHAIR evaluation, we select 100 images from the MSCOCO subset used in our CHAIR experiments and ask each method to generate free-form descriptions of the images using the same decoding setting as in \Secref{sec:results}. All methods are evaluated on a single GPU. We measure decoding throughput in tokens per second, which reflects the effective per-token computational cost of each method. \tabref{tab:runtime} shows the results.

\begin{table}[!t]
\centering
\setlength{\tabcolsep}{3pt}
\renewcommand{\arraystretch}{1.2}
\caption{Decoding throughput of different methods on 100 MSCOCO images used in the CHAIR evaluation. We report the number of tokens per second during generation (higher is better).}
\label{tab:runtime}
\vskip -0.2cm
\begin{tabular*}{1\linewidth}{l@{\extracolsep{\fill}}ccccccccc}
\toprule
\textbf{Method} & Base & COAD (Ours) & OPERA & PAI & DoLa & VCD & CAD & EOS & HALC \\
\midrule
\#tokens/s $\uparrow$ & 24.37 & 10.49 & 4.52 & 43.62 & 29.38 & 7.98 & 7.92 & 9.91 & 7.32 \\
\bottomrule
\end{tabular*}
\vskip -0.2cm
\end{table}

\textbf{Detector Overhead.}
\method{} invokes the object detector only once per image before decoding. Since autoregressive decoding dominates the total computational cost for MLLMs, this one-time detection overhead is relatively small: the RTMDet detector used in our implementation processes each image in about 0.10 seconds, which is negligible compared with the cumulative cost of token generation on long outputs.

\textbf{Dual-Model Decoding.}
\method{} evaluates both the pretrained model and the object-aware finetuned model at each decoding step. When executed sequentially on a single GPU, this dual-model decoding results in roughly half the throughput of the base LLaVA model (10.49 vs.~24.37 tokens/s in \tabref{tab:runtime}). However, the two forward passes are independent and can be executed fully in parallel on different GPUs, allowing throughput close to that of single-model decoding in practical multi-GPU deployments.

Compared to other hallucination-mitigation methods, \method{} remains computationally competitive. In particular, it is significantly faster than multi-step refinement and beam-search/backtracking approaches such as OPERA (4.52 tokens/s), and comparable to other decoding-modification methods like VCD, CAD, EOS, and HALC. Overall, \method{} achieves strong reductions in hallucination with a moderate and well-characterized runtime overhead.

\section{Conclusion}\label{sec:conclusion}

In this paper, we propose \method{}, a novel approach to reducing object hallucination in MLLMs. By combining object detection and causal inference, \method{} improves the quality of generated captions and reasoning outputs. Extensive experiments on various benchmarks show that \method{} consistently outperforms state-of-the-art dehallucination methods across diverse metrics and settings.

Future work may include more sophisticated object representations and extend our causal modeling framework to additional multimodal tasks. Since the
number of object categories is determined by the detector, it would also be
interesting to explore open-vocabulary detectors (e.g., GLIP \citep{glip}), which may allow
COAD to operate over a substantially richer and more flexible object space.
Moreover, we plan to investigate the integration of temporal and spatial priors to further enhance the causal grounding of visual elements. Another promising direction is to incorporate user feedback or human-in-the-loop supervision to dynamically refine the intervention policy during inference. Finally, we aim to explore the scalability of \method{} in real-world applications such as assistive vision systems and visually grounded dialogue.

In terms of limitations, like many other MLLMs, maliciously manipulated inputs could affect COAD’s performance. Another limitation is that COAD primarily targets object hallucination; extending the causal modeling framework to other forms of hallucination, such as attribute, relational, and global-scene inconsistencies, remains an important direction for future work. We defer a detailed discussion of limitations and potential mitigations to \appref{app:limitation}.

\bibliography{main}

@misc{llava,
      title={Improved Baselines with Visual Instruction Tuning}, 
      author={Haotian Liu and Chunyuan Li and Yuheng Li and Yong Jae Lee},
      year={2024},
      eprint={2310.03744},
      archivePrefix={arXiv},
      primaryClass={cs.CV},
      url={https://arxiv.org/abs/2310.03744}, 
}

@misc{cgd,
      title={Seeing is Believing: Mitigating Hallucination in Large Vision-Language Models via CLIP-Guided Decoding}, 
      author={Ailin Deng and Zhirui Chen and Bryan Hooi},
      year={2024},
      eprint={2402.15300},
      archivePrefix={arXiv},
      primaryClass={cs.CV},
      url={https://arxiv.org/abs/2402.15300}, 
}

@inproceedings{ICL,
  title={Causal discovery from incomplete data: A deep learning approach},
  author={Wang, Yuhao and Menkovski, Vlado and Wang, Hao and Du, Xin and Pechenizkiy, Mykola},
  booktitle={AAAI StarAI Workshop},
  year={2020}
}

@article{BDL,
  title={Towards Bayesian deep learning: A framework and some existing methods},
  author={Wang, Hao and Yeung, Dit-Yan},
  journal={TDKE},
  volume={28},
  number={12},
  pages={3395--3408},
  year={2016}
}

@article{BDLSurvey,
  title={A Survey on Bayesian Deep Learning},
  author={Wang, Hao and Yeung, Dit-Yan},
  journal={CSUR},
  volume={53},
  number={5},
  pages={1--37},
  year={2020}
}

@inproceedings{CounTS,
  author    = {Jingquan Yan and
               Hao Wang},
  title     = {Self-Interpretable Time Series Prediction with Counterfactual Explanations},
  booktitle = {ICML},
  year      = {2023}
}

@inproceedings{BLoB,
  title={BLoB: Bayesian Low-Rank Adaptation by Backpropagation for Large Language Models},
  author={Wang, Yibin and Shi, Haizhou and Han, Ligong and Metaxas, Dimitris and Wang, Hao},
  booktitle={NeurIPS},
  year={2024}
}

@misc{skipn,
      title={Skip \n: A Simple Method to Reduce Hallucination in Large Vision-Language Models}, 
      author={Zongbo Han and Zechen Bai and Haiyang Mei and Qianli Xu and Changqing Zhang and Mike Zheng Shou},
      year={2024},
      eprint={2402.01345},
      archivePrefix={arXiv},
      primaryClass={cs.CV},
      url={https://arxiv.org/abs/2402.01345}, 
}

@misc{cad,
      title={Trusting Your Evidence: Hallucinate Less with Context-aware Decoding}, 
      author={Weijia Shi and Xiaochuang Han and Mike Lewis and Yulia Tsvetkov and Luke Zettlemoyer and Scott Wen-tau Yih},
      year={2023},
      eprint={2305.14739},
      archivePrefix={arXiv},
      primaryClass={cs.CL},
      url={https://arxiv.org/abs/2305.14739}, 
}

@misc{vcd,
      title={Mitigating Object Hallucinations in Large Vision-Language Models through Visual Contrastive Decoding}, 
      author={Sicong Leng and Hang Zhang and Guanzheng Chen and Xin Li and Shijian Lu and Chunyan Miao and Lidong Bing},
      year={2023},
      eprint={2311.16922},
      archivePrefix={arXiv},
      primaryClass={cs.CV},
      url={https://arxiv.org/abs/2311.16922}, 
}

@misc{opera,
      title={OPERA: Alleviating Hallucination in Multi-Modal Large Language Models via Over-Trust Penalty and Retrospection-Allocation}, 
      author={Qidong Huang and Xiaoyi Dong and Pan Zhang and Bin Wang and Conghui He and Jiaqi Wang and Dahua Lin and Weiming Zhang and Nenghai Yu},
      year={2024},
      eprint={2311.17911},
      archivePrefix={arXiv},
      primaryClass={cs.CV},
      url={https://arxiv.org/abs/2311.17911}, 
}

@misc{pai,
      title={Paying More Attention to Image: A Training-Free Method for Alleviating Hallucination in LVLMs}, 
      author={Shi Liu and Kecheng Zheng and Wei Chen},
      year={2024},
      eprint={2407.21771},
      archivePrefix={arXiv},
      primaryClass={cs.CV},
      url={https://arxiv.org/abs/2407.21771}, 
}

@misc{eos,
      title={Less is More: Mitigating Multimodal Hallucination from an EOS Decision Perspective}, 
      author={Zihao Yue and Liang Zhang and Qin Jin},
      year={2024},
      eprint={2402.14545},
      archivePrefix={arXiv},
      primaryClass={cs.CL},
      url={https://arxiv.org/abs/2402.14545}, 
}

@misc{dola,
      title={DoLa: Decoding by Contrasting Layers Improves Factuality in Large Language Models}, 
      author={Yung-Sung Chuang and Yujia Xie and Hongyin Luo and Yoon Kim and James Glass and Pengcheng He},
      year={2024},
      eprint={2309.03883},
      archivePrefix={arXiv},
      primaryClass={cs.CL},
      url={https://arxiv.org/abs/2309.03883}, 
}

@book{pearl2009causality,
  title={Causality},
  author={Pearl, Judea},
  year={2009},
  publisher={Cambridge university press}
}

@misc{freshllms,
      title={FreshLLMs: Refreshing Large Language Models with Search Engine Augmentation}, 
      author={Tu Vu and Mohit Iyyer and Xuezhi Wang and Noah Constant and Jerry Wei and Jason Wei and Chris Tar and Yun-Hsuan Sung and Denny Zhou and Quoc Le and Thang Luong},
      year={2023},
      eprint={2310.03214},
      archivePrefix={arXiv},
      primaryClass={cs.CL},
      url={https://arxiv.org/abs/2310.03214}, 
}

@article{sun2023mmhal,
  title={Aligning large multimodal models with factually augmented rlhf},
  author={Sun, Zhiqing and Shen, Sheng and Cao, Shengcao and Liu, Haotian and Li, Chunyuan and Shen, Yikang and Gan, Chuang and Gui, Liang-Yan and Wang, Yu-Xiong and Yang, Yiming and others},
  journal={arXiv preprint arXiv:2309.14525},
  year={2023}
}

@article{rohrbach2018chair,
  title={Object hallucination in image captioning},
  author={Rohrbach, Anna and Hendricks, Lisa Anne and Burns, Kaylee and Darrell, Trevor and Saenko, Kate},
  journal={arXiv preprint arXiv:1809.02156},
  year={2018}
}

@article{li2023pope,
  title={Evaluating object hallucination in large vision-language models},
  author={Li, Yifan and Du, Yifan and Zhou, Kun and Wang, Jinpeng and Zhao, Wayne Xin and Wen, Ji-Rong},
  journal={arXiv preprint arXiv:2305.10355},
  year={2023}
}

@misc{stitchintime,
      title={A Stitch in Time Saves Nine: Detecting and Mitigating Hallucinations of LLMs by Validating Low-Confidence Generation}, 
      author={Neeraj Varshney and Wenlin Yao and Hongming Zhang and Jianshu Chen and Dong Yu},
      year={2023},
      eprint={2307.03987},
      archivePrefix={arXiv},
      primaryClass={cs.CL},
      url={https://arxiv.org/abs/2307.03987}, 
}

@misc{rarr,
      title={RARR: Researching and Revising What Language Models Say, Using Language Models}, 
      author={Luyu Gao and Zhuyun Dai and Panupong Pasupat and Anthony Chen and Arun Tejasvi Chaganty and Yicheng Fan and Vincent Y. Zhao and Ni Lao and Hongrae Lee and Da-Cheng Juan and Kelvin Guu},
      year={2023},
      eprint={2210.08726},
      archivePrefix={arXiv},
      primaryClass={cs.CL},
      url={https://arxiv.org/abs/2210.08726}, 
}

@misc{dress,
      title={DRESS: Instructing Large Vision-Language Models to Align and Interact with Humans via Natural Language Feedback}, 
      author={Yangyi Chen and Karan Sikka and Michael Cogswell and Heng Ji and Ajay Divakaran},
      year={2024},
      eprint={2311.10081},
      archivePrefix={arXiv},
      primaryClass={cs.CV},
      url={https://arxiv.org/abs/2311.10081}, 
}

@misc{recaption,
      title={Mitigating Fine-Grained Hallucination by Fine-Tuning Large Vision-Language Models with Caption Rewrites}, 
      author={Lei Wang and Jiabang He and Shenshen Li and Ning Liu and Ee-Peng Lim},
      year={2023},
      eprint={2312.01701},
      archivePrefix={arXiv},
      primaryClass={cs.CV},
      url={https://arxiv.org/abs/2312.01701}, 
}

@misc{hallucidoctor,
      title={HalluciDoctor: Mitigating Hallucinatory Toxicity in Visual Instruction Data}, 
      author={Qifan Yu and Juncheng Li and Longhui Wei and Liang Pang and Wentao Ye and Bosheng Qin and Siliang Tang and Qi Tian and Yueting Zhuang},
      year={2024},
      eprint={2311.13614},
      archivePrefix={arXiv},
      primaryClass={cs.CV},
      url={https://arxiv.org/abs/2311.13614}, 
}

@misc{lrv-instruction,
      title={Mitigating Hallucination in Large Multi-Modal Models via Robust Instruction Tuning}, 
      author={Fuxiao Liu and Kevin Lin and Linjie Li and Jianfeng Wang and Yaser Yacoob and Lijuan Wang},
      year={2024},
      eprint={2306.14565},
      archivePrefix={arXiv},
      primaryClass={cs.CV},
      url={https://arxiv.org/abs/2306.14565}, 
}

@misc{lvlm-hall-survey,
      title={A Survey on Hallucination in Large Vision-Language Models}, 
      author={Hanchao Liu and Wenyuan Xue and Yifei Chen and Dapeng Chen and Xiutian Zhao and Ke Wang and Liping Hou and Rongjun Li and Wei Peng},
      year={2024},
      eprint={2402.00253},
      archivePrefix={arXiv},
      primaryClass={cs.CV},
      url={https://arxiv.org/abs/2402.00253}, 
}

@article{llm-hall-survey,
   title={A Survey on Hallucination in Large Language Models: Principles, Taxonomy, Challenges, and Open Questions},
   ISSN={1558-2868},
   url={http://dx.doi.org/10.1145/3703155},
   DOI={10.1145/3703155},
   journal={ACM Transactions on Information Systems},
   publisher={Association for Computing Machinery (ACM)},
   author={Huang, Lei and Yu, Weijiang and Ma, Weitao and Zhong, Weihong and Feng, Zhangyin and Wang, Haotian and Chen, Qianglong and Peng, Weihua and Feng, Xiaocheng and Qin, Bing and Liu, Ting},
   year={2024},
   month=nov }

@misc{llama,
      title={LLaMA: Open and Efficient Foundation Language Models}, 
      author={Hugo Touvron and Thibaut Lavril and Gautier Izacard and Xavier Martinet and Marie-Anne Lachaux and Timothée Lacroix and Baptiste Rozière and Naman Goyal and Eric Hambro and Faisal Azhar and Aurelien Rodriguez and Armand Joulin and Edouard Grave and Guillaume Lample},
      year={2023},
      eprint={2302.13971},
      archivePrefix={arXiv},
      primaryClass={cs.CL},
      url={https://arxiv.org/abs/2302.13971}, 
}

@misc{minigpt,
      title={MiniGPT-4: Enhancing Vision-Language Understanding with Advanced Large Language Models}, 
      author={Deyao Zhu and Jun Chen and Xiaoqian Shen and Xiang Li and Mohamed Elhoseiny},
      year={2023},
      eprint={2304.10592},
      archivePrefix={arXiv},
      primaryClass={cs.CV},
      url={https://arxiv.org/abs/2304.10592}, 
}

@article{gpt4,
  title={Gpt-4 technical report},
  author={Achiam, Josh and Adler, Steven and Agarwal, Sandhini and Ahmad, Lama and Akkaya, Ilge and Aleman, Florencia Leoni and Almeida, Diogo and Altenschmidt, Janko and Altman, Sam and Anadkat, Shyamal and others},
  journal={arXiv preprint arXiv:2303.08774},
  year={2023}
}

@inproceedings{mscoco,
  title={Microsoft coco: Common objects in context},
  author={Lin, Tsung-Yi and Maire, Michael and Belongie, Serge and Hays, James and Perona, Pietro and Ramanan, Deva and Doll{\'a}r, Piotr and Zitnick, C Lawrence},
  booktitle={Computer vision--ECCV 2014: 13th European conference, zurich, Switzerland, September 6-12, 2014, proceedings, part v 13},
  pages={740--755},
  year={2014},
  organization={Springer}
}

@misc{rtmdet,
      title={RTMDet: An Empirical Study of Designing Real-Time Object Detectors}, 
      author={Chengqi Lyu and Wenwei Zhang and Haian Huang and Yue Zhou and Yudong Wang and Yanyi Liu and Shilong Zhang and Kai Chen},
      year={2022},
      eprint={2212.07784},
      archivePrefix={arXiv},
      primaryClass={cs.CV},
      url={https://arxiv.org/abs/2212.07784}, 
}

@article{halc,
  title={Halc: Object hallucination reduction via adaptive focal-contrast decoding},
  author={Chen, Zhaorun and Zhao, Zhuokai and Luo, Hongyin and Yao, Huaxiu and Li, Bo and Zhou, Jiawei},
  journal={arXiv preprint arXiv:2403.00425},
  year={2024}
}

@inproceedings{geninterventionforcausal,
  title={Generative interventions for causal learning},
  author={Mao, Chengzhi and Cha, Augustine and Gupta, Amogh and Wang, Hao and Yang, Junfeng and Vondrick, Carl},
  booktitle={Proceedings of the IEEE/CVF Conference on Computer Vision and Pattern Recognition},
  pages={3947--3956},
  year={2021}
}

@inproceedings{causaltransforvr,
  title={Causal transportability for visual recognition},
  author={Mao, Chengzhi and Xia, Kevin and Wang, James and Wang, Hao and Yang, Junfeng and Bareinboim, Elias and Vondrick, Carl},
  booktitle={Proceedings of the IEEE/CVF Conference on Computer Vision and Pattern Recognition},
  pages={7521--7531},
  year={2022}
}

@inproceedings{glip,
  title={Grounded language-image pre-training},
  author={Li, Liunian Harold and Zhang, Pengchuan and Zhang, Haotian and Yang, Jianwei and Li, Chunyuan and Zhong, Yiwu and Wang, Lijuan and Yuan, Lu and Zhang, Lei and Hwang, Jenq-Neng and others},
  booktitle={Proceedings of the IEEE/CVF conference on computer vision and pattern recognition},
  pages={10965--10975},
  year={2022}
}

@article{clip_guided,
  title={Seeing is believing: Mitigating hallucination in large vision-language models via clip-guided decoding},
  author={Deng, Ailin and Chen, Zhirui and Hooi, Bryan},
  journal={arXiv preprint arXiv:2402.15300},
  year={2024}
}
\bibliographystyle{tmlr}

\appendix
\section{Details on Implementation and Causal Graphs}\label{app:implement}

\textbf{Implementation Details.} 
We finetune \method{} on a subset of MSCOCO images sourced from the LLaVA dataset. To enable the model to incorporate the auxiliary input $\z$, we introduce a two-layer MLP projector (with hidden size $256$) that maps $\z$ into the token embedding space, following LLaVA’s multimodal token integration approach. We employ LoRA ($r=128$, $\alpha=256$), a cosine learning rate schedule with an initial learning rate of $4\text{e}{-5}$, a batch size of 128, and train the model for 1 epoch. To mitigate the model’s dependence on prior context, we apply Gaussian noise ($\sigma=0.005$) to the embeddings of previous tokens with a probability of 0.5 during training. 
During inference, we use sampling by default with temperature 0.2 and a maximum of 512 output tokens.
In implementing~\Eqref{eq:final-formula}, we find that it is more effective to perform fusion in the logit space rather than in the probability space. Therefore, we replace $P(y_f|\S,\x,\z)$ and $P(y_p|\S,\x)$ with their corresponding logits before computing the fused output, which is subsequently converted back to the probability space via softmax.

All experiments were conducted on a single machine with 8 NVIDIA RTX A5000 GPUs (24GB each), an AMD EPYC 7282 16-Core Processor (64 threads), and 256GB RAM. Finetuning typically took around 16 hours per model. Caption generation on 5,000 images took between 30 minutes and 2 hours, depending on the generation length.

For evaluation, we use sampling to generate outputs for all baseline methods, except for OPERA. Since OPERA is built on top of beam search, its outputs are generated using beam search with a beam size of 3 instead. For the hyperparameter $\alpha$ in \method{}, we set it to 1.5 for text generation tasks (CHAIR and MMHal-Bench) and 0.1 for POPE.

\textbf{Causal Graphs and Hidden States.} 
There are two different hidden states:
\begin{itemize}
    \item The \textbf{image-based, static hidden state $\z$}, which corresponds to the output of the MLLM's vision encoder (i.e., $H_v$ that we will further explain in Appendix~\ref{app:llava-arch}). In a causal MLLM such as LLaVA, the input follows the structure \textbf{``image | user prompt | generated tokens''}, and the hidden state of a token position is affected only by tokens that appear \textbf{before} it. Therefore, here $\z$ is determined solely by the image and aligns most closely with our interpretation of the object belief variable $\z$ in our causal graphs in \Figref{fig:temporal_causal_graphs} and \Figref{fig:causal-both}. \emph{It is not influenced by $\x$.}

    \item The \textbf{response-based, dynamic hidden state $\h$ (or $\h^{(t)}$)}, which corresponds to the hidden states when the MLLM generates the $t$-th response token. These hidden states are influenced by $\x$.
\end{itemize}

\begin{figure}[t]
  \centering
  \hfill
  \begin{subfigure}[t]{0.45\linewidth}
      \centering
      \includegraphics[width=0.7\linewidth]{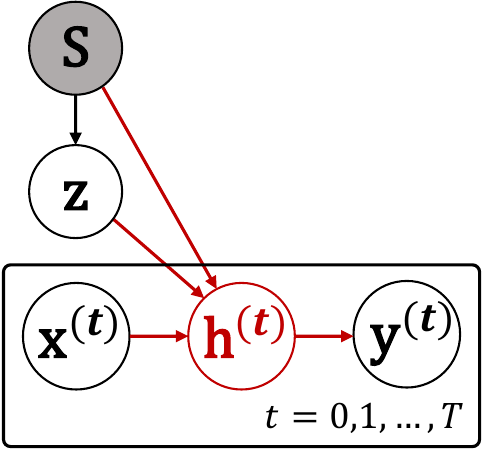}
      \caption{Casual graph with hidden states $\h^{(t)}$. The hidden states that decide $y^{(t)}$ are affected by $\S$, $\z$, and $\x^{(t)}$.}
      \label{fig:transient-a}
  \end{subfigure}
  \hfill
  \begin{subfigure}[t]{0.45\linewidth}
      \centering
      \includegraphics[width=0.7\linewidth]{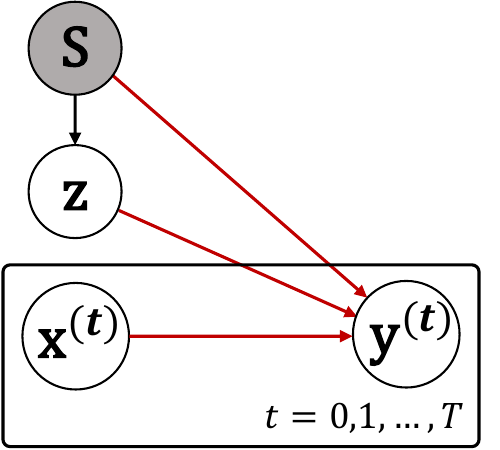}
      \caption{Collapsing hidden states $\h^{(t)}$ into the generation of $y^{(t)}$. The resulting graph corresponds to the causal graph of \method{} in \Figref{fig:causal-both}.}
      \label{fig:transient-b}
  \end{subfigure}
  \hfill
  \caption{
\textbf{The role of hidden states in the decoding causal graph of an MLLM.}
This figure illustrates a causal graph that explicitly includes token-level hidden states $\h^{(t)}$ during generation \textbf{(a)}, and shows that collapsing these hidden states into the generation of $y^{(t)}$ yields an equivalent causal graph consistent with the one used by \method{} \textbf{(b)}.
}
  \label{fig:transient}
\end{figure}

We further clarify the role of the dynamic hidden state $\h^{(t)}$ in the causal graph shown in \Figref{fig:transient}. In \Figref{fig:transient-a}:
\begin{itemize}
    \item the image $\S$ generates the image-based, static hidden state $\z$,
    \item $\S$, $\z$, and $\x^{(t)}$ then jointly generate (influence) the response-based, dynamic hidden state $\h^{(t)}$, and
    \item $\h^{(t)}$ then influences the generated token $y^{(t)}$.
\end{itemize}

As shown in \Figref{fig:transient-b}, we can actually collapse $h^{(t)}$ and the \emph{red} part of \Figref{fig:transient-a} to have an equivalent causal graph in \Figref{fig:transient-b}, which matches our causal graphs in \Figref{fig:temporal_causal_graphs} and \Figref{fig:causal-both}.

\section{Evidence of Hallucinatory Object Beliefs in LLaVA}
\label{app:evidence}

To examine whether a multimodal LLM (MLLM) can form \textit{incorrect internal beliefs} about object existence (corresponding to an incorrect $\z$ in our formulation), we conduct a linear-probe experiment on LLaVA. The goal is to determine whether LLaVA forms internal object-existence beliefs that deviate from the actual content of the image.

\subsection{Experimental Setup}

We train a linear classifier to probe whether LLaVA internally believes each object category to be present in the image. The classifier
\begin{itemize}
    \item takes as \textbf{input} the embedding for each token, and
    \item outputs a $C$-dimensional vector $\oo \in [0,1]^C$ indicating the estimated probability that each of the $C$ objects exists in the image.
\end{itemize}

Here, $\oo$ reflects the MLLM's object-existence belief in its hidden states, which we mentioned in Appendix~\ref{app:implement}.

\textbf{Input to the Linear Probe.}
For each token in the dialog sequence, LLaVA produces a hidden state after every transformer decoder layer. We concatenate these hidden states together with the embedding before the first layer to obtain the probe input. Thus, each token in the prompt-response sequence contributes one training sample for the linear probe.

\textbf{Conversation Sampling.}
We collect dialogs generated by LLaVA using MSCOCO images and the prompt ``Please describe every object in this image in detail.'' After LLaVA completes the response, we extract hidden states for all tokens (including image tokens, query tokens, and generated response tokens) to construct probe input samples.

\textbf{Target Labels.}
For each image, we run LLaVA once with the same prompt and use a rule-based method to determine whether each object category is mentioned in the resulting text description. These detected mentions serve as labels for training the linear probe. The same label is assigned to all probe samples within a given image-prompt-response tuple.

\textbf{Discussion on the Relation between Input to the Linear Probe and the Hidden States.}
In a causal MLLM such as LLaVA, the input follows the structure \textbf{``image | user prompt | generated tokens''}, and the hidden state of a token position is affected only by tokens that appear \textbf{before} it. Therefore:
\begin{itemize}
    \item when the input to the linear probe is the hidden state in the ``image'' part, the output $\oo$ serves as an estimation of the \textbf{image-based, static hidden state $\z$};
    \item when the input to the linear probe is the hidden state in the ``generated tokens'' part, the output $\oo$ serves as an estimation of the \textbf{response-based, dynamic hidden state $\h$ (or $\h^{(t)}$)}.
\end{itemize}

\begin{figure}[!p]
\vskip -1cm
    \centering

    \begin{minipage}[c]{0.73\linewidth}
        \centering
        \vfill
        \includegraphics[width=\linewidth]{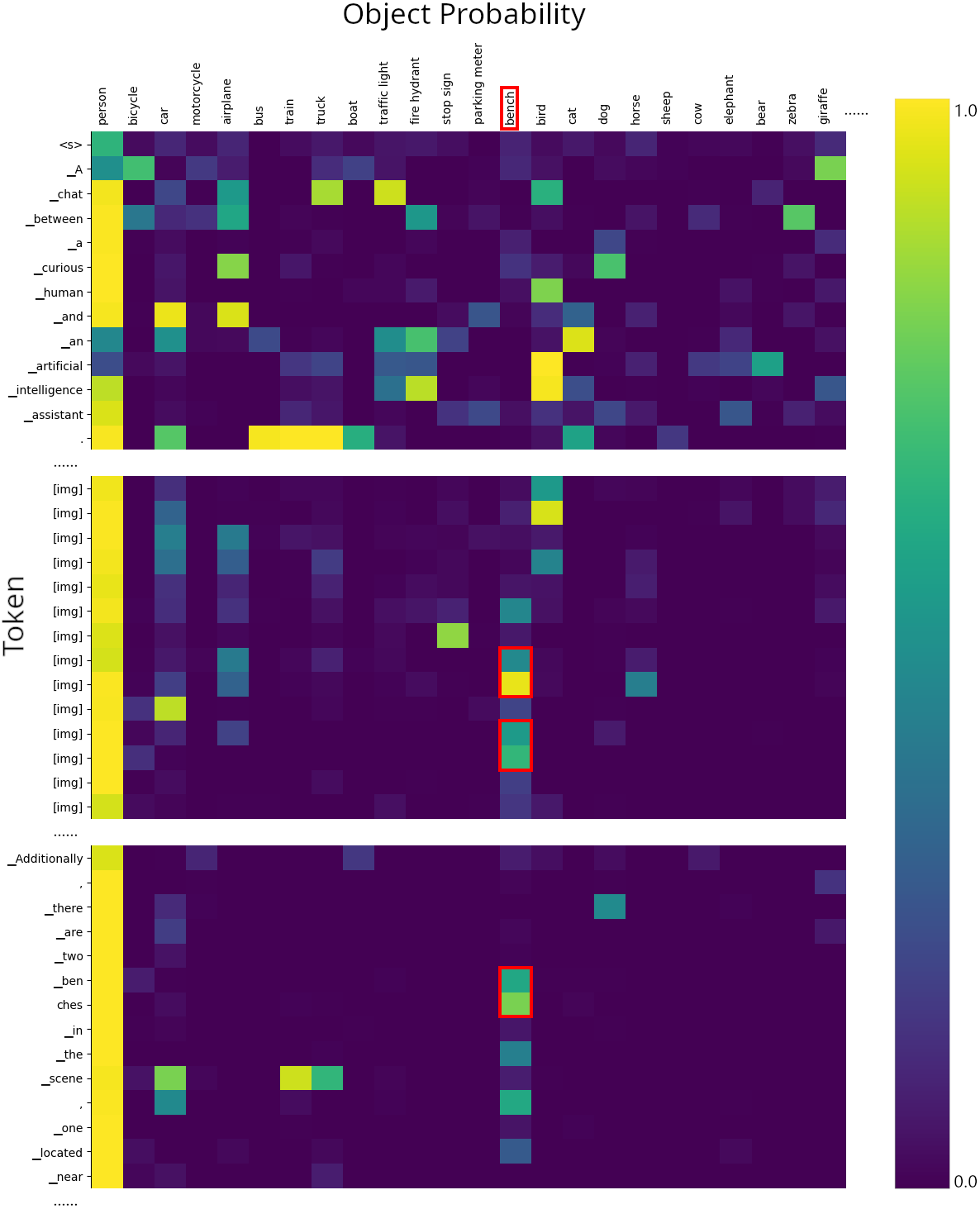}
        \vskip -0.4cm
        \caption*{(a) Linear-probe output $\oo$ for object probabilities in LLaVA.}
        \vfill
    \end{minipage}
        \centering
        \vspace{0.2cm}
        \vfill
        \begin{minipage}[t]{0.27\linewidth}
            \centering
            \includegraphics[height=3.6cm]{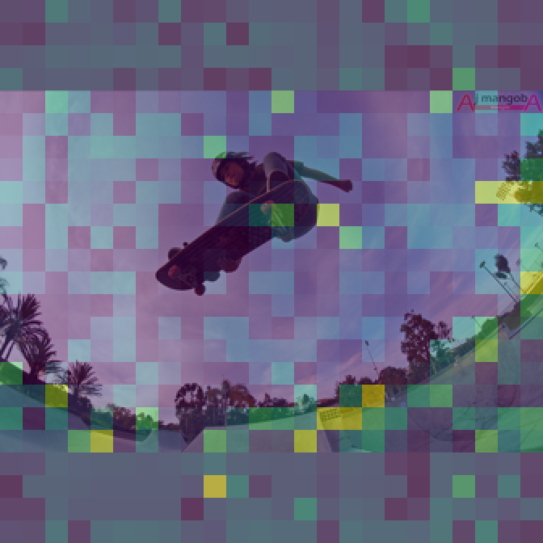}
            \vskip -0.2cm
            \caption*{(b) Visualization of probed probability of the object \textit{``bench''}  over all image tokens.}
        \end{minipage}
        \hfill
        \begin{minipage}[t]{0.34\linewidth}
            \centering
            \includegraphics[width=\linewidth]{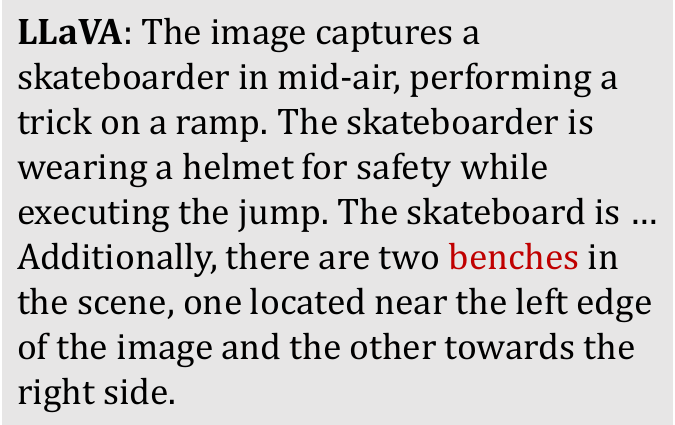}
            \vskip -0.2cm
            \caption*{(c) LLaVA's response hallucinates the object ``\textit{benches}''.}
        \end{minipage}
        \hfill
        \begin{minipage}[t]{0.34\linewidth}
            \centering
            \includegraphics[width=\linewidth]{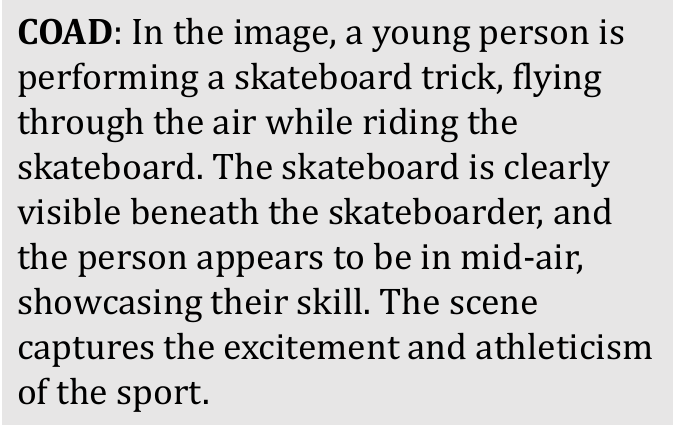}
            \vskip -0.2cm
            \caption*{(d) \method{} eliminates the bench-related hallucination.}
        \end{minipage}

\vskip -0.2cm
    \caption{
        \textbf{Visualization of LLaVA's hallucinatory object beliefs.} 
This figure provides empirical evidence that LLaVA can form incorrect internal object-existence beliefs (corresponding to an incorrect $\z$) and shows how \method{} corrects the resulting hallucination.
\textbf{(a)} Linear-probe output of LLaVA's object probabilities for all tokens over the full dialog, showing that the probed probability for the object \textit{``bench''} increases both around \emph{certain image tokens} and around the \emph{tokens where LLaVA actually generates the word ``\textit{benches}''} (in \red{red} boxes).
\textbf{(b)} The baseline LLaVA's probed probability of the object \textit{``bench''} over the image tokens (patches), showing high \textit{``bench''} probabilities in some image tokens (highlighted in yellow); this indicates that the hallucinated belief may originate from the image-perception stage. 
\textbf{(c)} LLaVA hallucinates the object \textit{``bench''} in the response. 
\textbf{(d)} \method{}’s response, which eliminates the bench hallucination.
    }
    \label{fig:evidence}
\end{figure}

\subsection{Findings}

We apply the trained linear probe to a sampled conversation and visualize the predicted object probabilities over time (\Figref{fig:evidence}).
\begin{itemize}
    \item The input image contains a person riding a skateboard with \textbf{no benches} in the background.
    \item The probe output in \Figref{fig:evidence}(a) indicates that LLaVA internally assigns a high probability (i.e., brighter, yellow color in the heatmap) to a nonexistent \textit{bench} around certain image-token positions.
    \item \Figref{fig:evidence}(b) shows LLaVA's probed probability of the object \textit{``bench''} over the image tokens (patches), demonstrating high \textit{``bench''} probabilities in some image tokens (highlighted in yellow); this indicates that the baseline LLaVA's hallucinated belief may originate from the image-perception stage. 
    \item \Figref{fig:evidence}(c) shows that LLaVA hallucinates the object \textit{``bench''} in the response. 
    \item In contrast, \method{} produces a response that does not mention any benches, as shown in \Figref{fig:evidence}(d).
\end{itemize}

\subsection{Conclusion}

This experiment demonstrates that LLaVA can form incorrect internal beliefs about object existence, i.e., it may estimate an incorrect $\z$ for certain objects, which subsequently leads to hallucinations. These results provide direct empirical evidence for the assumption illustrated in \Figref{fig:teaser} of the main paper.

\section{LLaVA Architecture and the Formation of Visual Features}
\label{app:llava-arch}

\begin{figure}[t]
    \centering
    \includegraphics[width=0.85\linewidth]{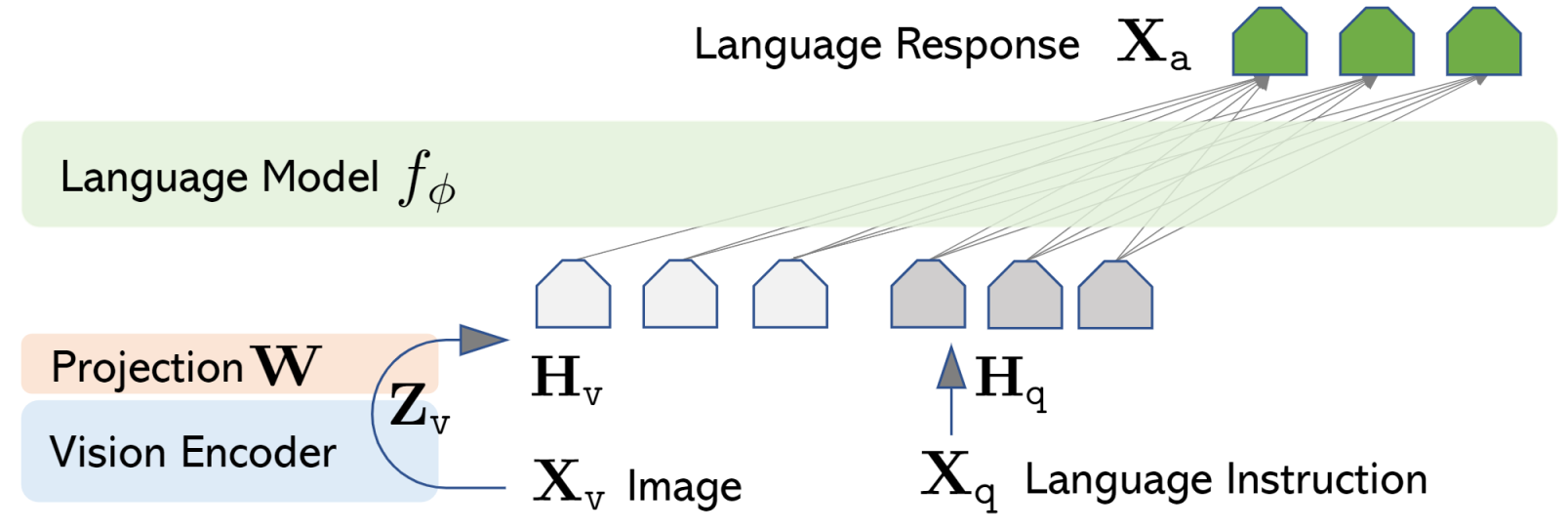}
    \caption{
        \textbf{Simplified LLaVA architecture \citep{llava}.}
        The image is encoded by a vision encoder and then projected into the
        language model’s embedding space. The resulting visual embeddings $H_v$
        are inserted as image tokens. Because $H_v$ is computed independently of
        textual tokens, the object-belief variable $\z$ corresponding to $H_v$
        remains fixed during decoding. (This figure is adapted from~\citep{llava}.)
    }
    \label{fig:llava-arch}
\end{figure}

To clarify why the object-belief variable $\z$ is treated as time-invariant in the causal model of a standard MLLM (\Secref{sec:mlm-causal-model}), we briefly describe the relevant components of the LLaVA architecture. A schematic diagram is shown in \Figref{fig:llava-arch}.

\subsection{Visual Feature Extraction Before Language Interaction}

In LLaVA, the input image is first processed by a vision encoder (e.g., CLIP’s
ViT backbone) to produce a set of image features, denoted $Z_v$. These features are then passed through a projector layer to obtain the final visual embeddings $H_v$, which are fed into the language model.

Importantly:
\begin{itemize}
    \item Both the vision encoder and the projector operate \emph{independently}
          of the language model.
    \item No textual tokens are involved when computing $Z_v$ or $H_v$.
    \item The visual embeddings $H_v$ remain fixed throughout decoding.
\end{itemize}

The projected embeddings $H_v$ are inserted into the token sequence as
``image tokens,'' and only after this insertion does the multimodal Transformer
start attending jointly over image tokens and text tokens.

\subsection{Relation to the Object-Belief Variable $\z$}

In our causal formulation, the variable $\z$ represents the model’s internal
belief about which objects exist in the image. Conceptually, $\z$ corresponds to
the portion of the visual embeddings $H_v$ that encode object presence or
absence. Since $H_v$ is generated solely from the image (via the vision encoder
and projector) and does not depend on previously generated text tokens, $\z$ is
naturally treated as a \textbf{static} variable. This matches the assumption used in
\Secref{sec:mlm-causal-model}, where only $\x^{(t)}$ and $y^{(t)}$ evolve
over time.

\section{Further Analysis of Ablation Studies}
\label{app:ablation}

\textbf{Effect of Finetuning and Effectiveness of Our Causal Decoding Algorithm.} {Since \method{} involves finetuning an MLLM, a natural question is whether the observed gains are simply due to finetuning rather than our proposed causal decoding algorithm. To examine this, we directly evaluate the finetuned model $M_f$ without applying our decoding strategy. As shown in \tabref{tab:chair}, $M_f$ alone achieves only part of the improvements, indicating that finetuning by itself cannot account for the performance of \method{} and verifying the effectiveness of our causal decoding algorithm.}

\textbf{Role of the Vector $\mathbf{z}$.} {Another question is whether the improvements come merely from contrasting $M_f$ and $M_p$ during causal fusion, regardless of our vector $\mathbf{z}$. To verify this, we remove $\mathbf{z}$ when training $M_f$ (i.e., a standard finetuning setting) and then apply our causal decoding procedure using this variant of $M_f$. The results from \tabref{tab:chair} show a clear drop compared to \method{}, demonstrating that $\mathbf{z}$ plays an essential role in enabling effective causal fusion.}

\section{Limitations and Future Work}
\label{app:limitation}

\textbf{Dependence on Finetuning.} 
Our approach currently requires a LoRA finetuning step for adaptation. While this is practical in many settings, it reduces the plug-and-play convenience of COAD. We experimented with a training-free variant that injects the causal vector directly as a prompt, which already yields strong improvements on POPE benchmarks (e.g., F1-Rand = 95.4, F1-Pop = 90.0, F1-Adv = 85.8), outperforming all baselines. However, this variant is more sensitive to detector errors and less robust on captioning tasks. These results nonetheless highlight the generality of COAD and suggest promising directions for reducing computational cost, such as improving detector reliability or designing dedicated training methods that allow a single MLLM to simulate the causal signal.

\textbf{Domain Mismatch.}
COAD relies on detectors trained on specific distributions. If the test image domain diverges significantly, the causal signal may become insufficient. One direction is to investigate zero-shot or domain-adaptive detectors to mitigate this issue.

\textbf{Adversarial Vulnerability.}
Maliciously manipulated inputs or detector outputs could affect COAD’s performance. However, its modular design allows for safeguard components (e.g., adversarial detection at the detector level), which we leave as future extensions.

\textbf{Residual Text Priors.}
In rare cases with extremely strong linguistic priors, causal interventions may not fully suppress hallucinations. In rare cases with extremely strong linguistic priors, causal interventions may not fully suppress hallucinations. Future improvements may involve designing stronger intervention mechanisms or complementary signals that better counteract such priors.

\textbf{Scope and Generality.}
Our study mainly targets object hallucinations. Broader validation on other types of hallucinations and across more MLLMs is a promising next step, facilitated by COAD’s general token-based interface.

\section{More Qualitative Examples}\label{app:case_study}

We provide more qualitative examples in \Figref{fig:case-studies1} and \Figref{fig:case-studies2}, where MSCOCO objects mentioned in the text are highlighted in \textcolor[RGB]{192,0,0}{red} (hallucinated) or \textcolor[RGB]{0,128,0}{green} (correct).

\newcommand{\zbox}[1]{%
  \colorbox{gray!10}{%
    \parbox{\dimexpr\linewidth-2\fboxsep\relax}{#1}%
  }%
}

\newcommand{\casecardZ}[4]{%
  \begin{minipage}[t]{0.48\linewidth}
    \makebox[\linewidth]{%
\includegraphics[width=\linewidth,height=\cardimgheight,keepaspectratio]{#1}%
}
    \vspace{3pt}

    \par
    \zbox{\textbf{$\z$-vector:} #2}\vspace{3pt}

    \fbox{\parbox{\dimexpr\linewidth-2\fboxsep-2\fboxrule\relax}{\textbf{LLaVA: } #3}}

    \vspace{1pt}

    \fbox{\parbox{\dimexpr\linewidth-2\fboxsep-2\fboxrule\relax}{\textbf{COAD: } #4}}
  \end{minipage}%
}

\newlength{\cardimgheight}
\setlength{\cardimgheight}{5cm}

\begin{figure}[h]
\centering
\begingroup
  \setkeys{Gin}{height=\cardimgheight,keepaspectratio}

  \casecardZ{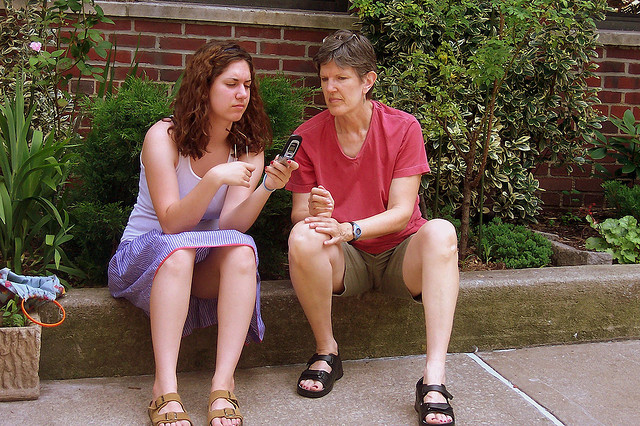}{[person: 0.93, handbag: 0.76, cell phone: 0.72, potted plant: 0.5, clock: 0.26, bottle: 0.24, \textbf{bench: 0.2}, ...]}{A \textcolor[RGB]{0,128,0}{woman} is sitting on a \textcolor[RGB]{192,0,0}{bench}, holding a \textcolor[RGB]{0,128,0}{cell phone} and talking to another woman. They are both wearing sandals.}{A \textcolor[RGB]{0,128,0}{woman} in a pink shirt holds a \textcolor[RGB]{0,128,0}{cell phone}, while another \textcolor[RGB]{0,128,0}{woman} in a dress sits next to her.}
  \hfill
  \casecardZ{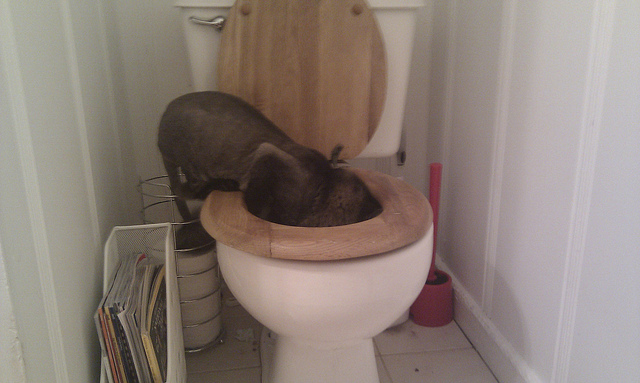}{[toilet: 0.86, cat: 0.84, book: 0.46, \textbf{dog: 0.17}, ...]}{A \textcolor[RGB]{192,0,0}{dog} is standing on a \textcolor[RGB]{0,128,0}{toilet}, sniffing the \textcolor[RGB]{0,128,0}{toilet} bowl. The \textcolor[RGB]{0,128,0}{toilet} lid is open, and the \textcolor[RGB]{192,0,0}{dog} appears to be curious about the water inside.}{A \textcolor[RGB]{0,128,0}{cat} is sitting on the \textcolor[RGB]{0,128,0}{toilet} and appears to be drinking water from the \textcolor[RGB]{0,128,0}{toilet} \textcolor[RGB]{0,128,0}{bowl}.}

  \vspace{6pt}

  \casecardZ{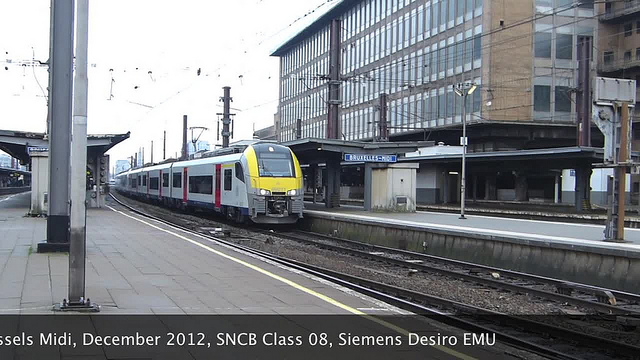}{[train: 0.91, chair: 0.26, bird: 0.21, \textbf{bench: 0.2}, person: 0.16, clock: 0.15, traffic light: 0.14, ...]}{A yellow and white \textcolor[RGB]{0,128,0}{train} is traveling down the tracks, pulling into a station. The \textcolor[RGB]{0,128,0}{train} is positioned next to a building, and there is a \textcolor[RGB]{192,0,0}{bench} nearby for \textcolor[RGB]{192,0,0}{passenger} to sit. The scene is captured on November 1, 2012.}{A \textcolor[RGB]{0,128,0}{train} is stopped at a \textcolor[RGB]{0,128,0}{train} station, and there is a yellow line on the platform.}
  \hfill
  \casecardZ{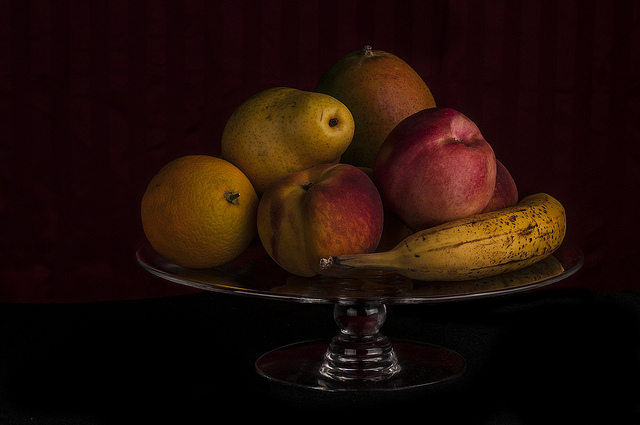}{[banana: 0.87, apple: 0.87, orange: 0.83, bowl: 0.33, \textbf{dining table: 0.28}, wine glass: 0.11, cup: 0.05, ...]}{A clear glass plate is filled with a variety of fruits, including \textcolor[RGB]{0,128,0}{apple}, \textcolor[RGB]{0,128,0}{orange}, and a banana. The plate is placed on a \textcolor[RGB]{192,0,0}{table}, and the fruits are arranged in a visually appealing manner.}{A clear glass plate holds a variety of fruits, including \textcolor[RGB]{0,128,0}{banana}, \textcolor[RGB]{0,128,0}{apple}, pears, and an \textcolor[RGB]{0,128,0}{orange}.}
\endgroup

\caption{Case studies on hallucination reduction. }
\label{fig:case-studies1}
\end{figure}


\begin{figure}[h]
\centering
\begingroup
  \setkeys{Gin}{height=\cardimgheight,keepaspectratio}

  \casecardZ{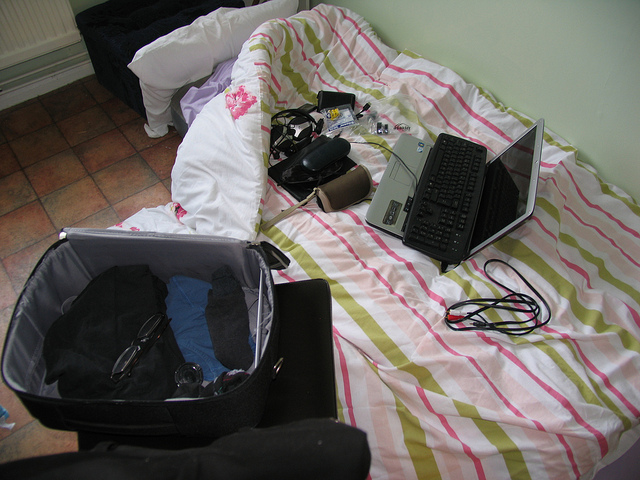}{[laptop: 0.88, bed: 0.86, suitcase: 0.63, keyboard: 0.44, mouse: 0.42, backpack: 0.36, handbag: 0.33, \textbf{cell phone: 0.31}, remote: 0.15, person: 0.14, ...]}{A \textcolor[RGB]{0,128,0}{bed} with a \textcolor[RGB]{0,128,0}{laptop} and a \textcolor[RGB]{192,0,0}{cell phone} on it. The \textcolor[RGB]{0,128,0}{bed} is unmade and has a striped blanket. There is also a \textcolor[RGB]{0,128,0}{suitcase} on the \textcolor[RGB]{0,128,0}{bed}, and a pair of glasses nearby.}{A \textcolor[RGB]{0,128,0}{laptop} is sitting on top of a \textcolor[RGB]{0,128,0}{bed} next to a \textcolor[RGB]{0,128,0}{suitcase}.}
    \hfill
    \casecardZ{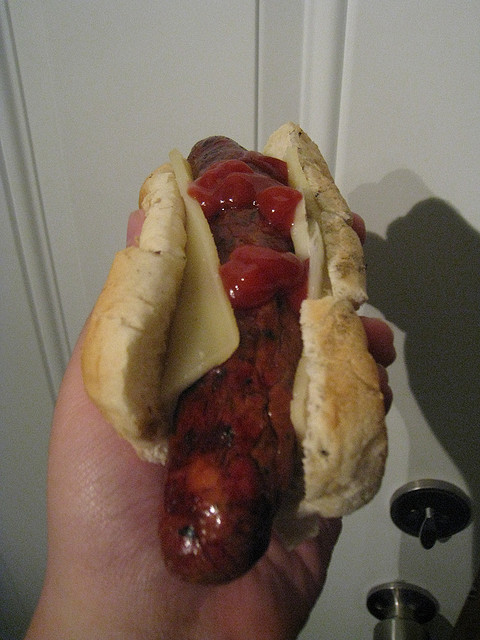}{[hot dog: 0.91, person: 0.77, sandwich: 0.31, bottle: 0.16, pizza: 0.07, ..., \textbf{dining table: 0.04}, ...]}{A \textcolor[RGB]{0,128,0}{person} is holding a \textcolor[RGB]{0,128,0}{hot dog} with cheese and ketchup on a bun. The \textcolor[RGB]{0,128,0}{hot dog} is placed on a \textcolor[RGB]{192,0,0}{table}, and the \textcolor[RGB]{0,128,0}{person}'s hand is visible in the foreground.}{A \textcolor[RGB]{0,128,0}{hot dog} with ketchup and cheese being held by a \textcolor[RGB]{0,128,0}{person}.}

  \vspace{6pt}

  \casecardZ{}{[bicycle: 0.89, person: 0.83, motorcycle: 0.75, backpack: 0.34, bottle: 0.27, handbag: 0.18, ..., \textbf{chair: 0.05}, ...]}{A \textcolor[RGB]{0,128,0}{man} is working on a \textcolor[RGB]{0,128,0}{motorcycle} in a garage, standing on a red \textcolor[RGB]{192,0,0}{stool} to reach the motorcycle. He is wearing jeans and a black shirt. The \textcolor[RGB]{0,128,0}{motorcycle} is parked on a lift, and there are other \textcolor[RGB]{0,128,0}{motorcycle} in the background.}{A \textcolor[RGB]{0,128,0}{man} is working on a \textcolor[RGB]{0,128,0}{motorcycle}, lifting the front wheel off the ground using a lift.}
  \hfill
  \casecardZ{}{[person: 0.85, bird: 0.83, cup: 0.83, sandwich: 0.64, \textbf{bench: 0.55}, dining table: 0.5, fork: 0.33, chair: 0.32, knife: 0.22, ...]}{A \textcolor[RGB]{0,128,0}{bird} is standing on a plate with a half-eaten \textcolor[RGB]{0,128,0}{sandwich}, which is placed on a dining table. The \textcolor[RGB]{0,128,0}{bird} seems to be interested in the \textcolor[RGB]{0,128,0}{sandwich}, possibly trying to get a bite. The scene takes place near a body of water, with a \textcolor[RGB]{192,0,0}{bench} nearby.}{A half-eaten \textcolor[RGB]{0,128,0}{sandwich} sits on a plate with ketchup, and a \textcolor[RGB]{0,128,0}{bird} is standing nearby, possibly interested in the remaining food.}

\endgroup

\caption{Case studies on hallucination reduction. }
\label{fig:case-studies2}
\end{figure}

\end{document}